\pdfoutput=1

\documentclass[11pt]{article}

\usepackage{algorithm}
\usepackage[noend]{algpseudocode}

\usepackage{acl}

\usepackage{times}
\usepackage{latexsym}

\usepackage[T1]{fontenc}

\usepackage[utf8]{inputenc}

\usepackage{microtype}

\usepackage{inconsolata}

\usepackage{amsmath}
\usepackage{graphicx}
\usepackage{booktabs}
\usepackage{amsfonts}
\usepackage{multirow}
\usepackage{enumitem}
\usepackage{lscape}
\usepackage{arydshln}
\usepackage{tabularx}
\newcommand\setrow[1]{\gdef\rowmac{#1}#1\ignorespaces}
\newcommand\clearrow{\global\let\rowmac\relax}
\clearrow

\title{Reliability Estimation of News Media Sources: \textit{Birds of a Feather Flock Together}}

\author{Sergio Burdisso$^{\star,1}$, Dairazalia Sánchez-Cortés$^1$, Esaú Villatoro-Tello$^1$ \and Petr Motlicek$^{1,2}$ \\
         $^1$Idiap Research Institute, Martigny, Switzerland \\ $^2$Brno University of Technology, Brno, Czech Republic\\
         \texttt{\{sergio.burdisso,dairazalia.sanchez-cortes,esau.villatoro,petr.motlicek\}@idiap.ch}}

\begin{document}
\maketitle
\begin{abstract}

Evaluating the reliability of news sources is a routine task for journalists and organizations committed to acquiring and disseminating accurate information.
Recent research has shown that predicting sources' reliability represents an important first-prior step in addressing additional challenges such as fake news detection and fact-checking.
In this paper, we introduce a novel approach for source reliability estimation that leverages reinforcement learning strategies for estimating the reliability degree of news sources. Contrary to previous research, our proposed approach models the problem as the estimation of a reliability degree, and not a reliability label, based on how all the news media sources interact with each other on the Web.
We validated the effectiveness of our method on a news media reliability dataset that is an order of magnitude larger than comparable existing datasets. Results show that the estimated reliability degrees strongly correlates with journalists-provided scores (Spearman=0.80) and can effectively predict reliability labels (macro-avg. F$_1$ score=81.05).
We release our implementation and dataset, aiming to provide a valuable resource for the NLP community working on information verification.

\end{abstract}

\section{Introduction}

As of 2023, the number of internet users is over 5.18 billion worldwide, meaning that around two-thirds of the global population is currently connected to the WWW~\citep{statista2023}.
The Web has democratized and radically changed how people consume and produce information by shifting the paradigm from a news-centred one to a user-centred one.
Nowadays, any person on the Web can potentially be a ``news medium'' providing information either by creating websites, blogs and/or by making use of social media platforms.

Nevertheless, news media can no longer perform its role as “gatekeeper” deciding which stories to disseminate to the public or not~\citep{munger2020all} since most of the information on the Internet is unregulated by nature~\citep{cuan2020misinformation}.
As a consequence, an enormous proliferation of misinformation has emerged leaving the public vulnerable to incorrect or misleading information about the state of the world which, among others, increased polarization and decreased trust in institutions and experts~\citep{lewandowsky2017beyond, stromback2020news}.
The World Health Organization (WHO) recently declared a worldwide ``infodemic'' characterized by an overabundance of misinformation~\citep{van2022misinformation}.
The best-known type of misinformation is \textit{fake news}~\citep{lazer2018science} defined as ``false information intentionally created to mislead and/or manipulate a public through the appearance of a news format with an opportunistic structure to attract the reader’s attention''~\citep{baptista2022working}.

In an attempt to limit the impact of fake news, a large number of initiatives have been undertaken by media, journalists, governments, and international organizations to identify true and false information across the globe~\citep{shaar-etal-2020-known}.
For instance, the Duke University's center for journalism research, the Reporters’ Lab, lists a total of 419 fact-checking active sites online\footnote{\url{https://reporterslab.org/fact-checking/} (Oct. 2023)} from which FactCheck.org, Snopes, Full Fact and Politifact are the most well-known.
These sites manually and systematically assess the validity of thousands of claims. However, human annotators will always be outnumbered by the claims that need to be verified, reducing the impact of such services in a large-scale scenario. Consequently, we have witnessed a growing interest in using different machine learning models, ranging from non-neural~\citep{kwon2013aspects,popat2016credibility,nguyen2018interpretable} to deep learning-based ones~\citep{ma2016detecting, wang-2017-liar, popat-etal-2018-declare, wang2018eann, fajcik-etal-2023-claim}, to determine the validity of claims, news and online information. 
Nevertheless, these models' performance still has not reached confident accuracy values, limiting their applicability in real-world scenarios~\citep{baly-etal-2018-predicting}.

A more recent paradigm to fight fake news proposes to focus on the source rather than on the content~\citep{baly-etal-2018-predicting, baly-etal-2020-written}, a task referred to as profiling news medium.  The underlying hypothesis states that even though fake news spreads mainly through social media, they still need an initial website hosting the news. Hence, if information about websites is known in advance, identifying potentially fake news can be done by verifying the \textit{reliability} of the source.
In fact, this activity is also performed by journalists, who often consult rating services for news websites like NewsGuard\footnote{\url{https://www.newsguardtech.com/}} or MBFC\footnote{\url{https://mediabiasfactcheck.com}}.
Nonetheless, these services are not exhaustive and difficult to keep up-to-date as they rely on human evaluators, highlighting the need for scalable automatic solutions that can be applied in real-world scenarios.
Previous research has shown that predicting reliability is an important first-prior step for fact-checking systems~\citep{nguyen2018interpretable,popat2017truth,mukherjee2015leveraging} and also the most important aspect that journalists consider when manually verifying the trustworthiness of the information~\citep{baly-etal-2018-predicting}. Thus, in this paper, we focus on the task of \textit{source reliability estimation}, i.e., automatically analyzing the source that produces a given piece of information and determining its reliability degree. 
Concretely, we address the posed task by investigating the following research question: \textit{to what extent can we predict the reliability of a news media source solely based on its interactions with other sources?} Contrary to previous research, our proposed method represents a scalable and language-independent approach that can be further enriched via content-based features. Our performed experiments shed light on the immediate (positive) effects of profiling news mediums through its interactions with other sources and also in combination with traditional content-based attributes. Our research holds the potential to uncover deeper insights by incorporating more recent content-based technologies to further explore the nuanced dynamics of news websites, opening the door to a broader NLP research community.

Overall, the main contributions of this paper can be summarized as follows: \textit{(i)} we propose a methodology capable of modeling the \textit{source reliability estimation} problem in a real-world scale scenario that contrary to previous research, estimates the reliability degree (\textit{i.e.} a continuous value) rather than a categorical value and does not depend on any third-party resources; \textit{(ii)} we pioneer the introduction and evaluation of different algorithms to estimate the reliability score, exploring a spectrum from vanilla reinforcement learning strategies to task-specific variations; \textit{(iii)} we build the largest news media reliability dataset available, orders of magnitude larger than existing datasets; \textit{(iv)} we present empirical evidence that establishes the feasibility of predicting the reliability of a news media source solely through its interactions with other sources (which further improves when content-based features are incorporated); 
and \textit{(v)} we release both the dataset and source code to the wider NLP research community.\footnote{\url{https://github.com/idiap/News-Media-Reliability}}

\section{Related Work}
\label{sec:related-work}

The task of determining information veracity has been approached from different angles and perspectives, from micro to macro, depending on the object of study.
For instance, fact-checking focuses on validating a single statement, \textit{i.e.} the claim; fake news detection analyses a whole document, \textit{i.e.} the content of the news article.
In this work, we focus on the source that produces a given piece of information, also known as \textit{source reliability estimation}.

Within social media, the sources are individual users creating the content, and previous work has focused on identifying different types of  users such as \textit{spammers}~\citep{liubchenko2022research,stringhini2010detecting}, \textit{bots}~\citep{lei-etal-2023-bic, knauth-2019-language}, fake profiles~\citep{pradeep2020,ramalingam2018fake}, paid users~\citep{mihaylov-etal-2015-exposing}, and \textit{trolls}~\citep{fi12020031,miao-etal-2020-detecting,mihaylov-etal-2015-finding}, among others~\citep{sansonetti2020unreliable,burdisso-etal-2022-idiapers}.
However, in the broader case of the WWW, sources are individual websites~\citep{dong2015knowledge}, and in our case, news media websites.

Previous studies have tangentially addressed news media source reliability as part of the study of automatic fact-checking systems, either as a \textit{prior} in probabilistic graphical models~\citep{nguyen2018interpretable,mukherjee2015leveraging} or as features for stance classification models~\citep{popat2018credeye,popat2017truth,popat2016credibility}.
In these studies, reliability estimation relied on indirect measures since no gold labels were used.
For instance, some works use the \textit{AlexaRank}\footnote{\url{https://www.alexa.com/}} and \textit{PageRank}~\citep{brin1998107} scores of the websites as proxies for their reliability~\citep{baly-etal-2018-predicting,popat2016credibility} while others the proportion of articles that refute false claims and support true claims~\citep{popat2018credeye,popat2017truth}.
However, in the latter, authors rely on a fact-checking model and the selected true and false claims while, in the former, on scores that only capture the authority and popularity of the sources, not necessarily their trustworthiness --- for instance, think of popular unreliable gossip websites\footnote{\url{http://www.ebizmba.com/articles/gossip-websites}} or satirical news websites, like \textit{The Onion},\footnote{\url{https://www.theonion.com/}} highly popular, attracting huge web traffic.

Recently, ~\citet{baly-etal-2020-written,baly-etal-2019-multi,baly-etal-2018-predicting} addressed the source reliability estimation task on its own, modeling it as a classification task using source-level gold annotations.
In particular, authors focused on predicting websites factual reporting and political bias using the values published by a news media rating service as ground truth.
However, their proposed method relies on collecting and extracting information from multiple external and restricted sources (e.g. Twitter, Facebook, YouTube, etc.) for generating content-based, audience-based, and metadata-based features for the classification model, limiting its practical use on a large-scale scenario.
In this paper, we also address the task using gold annotations, however, we adopt an easier-to-scale approach.
Specifically, we model the problem as estimating a continuous value (i.e. the reliability \textit{degree}) based simply on how all news media sources interact with each other on the World Wide Web.

\section{Methodology}
\label{sec:method}


\subsection{Problem Formulation}
\label{subsec:problem-formulation}

Let $S$ be the set of all news media sources on the Web, and $G=\langle S, E, w\rangle$ be the weighted directed graph where there is an edge $(s, s') \in E$ if source $s$ contains articles (hyper) linked to $s'$ and where the weight $w(s, s') \in [0, 1]$ is the proportion of total hyperlinks in $s$ linked to $s'$.
Given two disjoint subsets $\hat{S}^+, \hat{S}^- \subset S$ containing, respectively, some known reliable and unreliable news sources, the goal is to estimate the \textit{reliability degree} $\rho(s)$ \textit{for all} $s\in S$.
More precisely, a total function $\rho:S \mapsto \mathbb{R}$ such that:
\begin{enumerate}[itemsep=1pt]
    \item $\rho(s) > 0$ if $s$ is reliable
    \item $\rho(s) \le 0$ if $s$ is unreliable
    \item $\rho(s) < \rho(s')$ if $s'$ is more reliable than $s$.
\end{enumerate}

The underlying intuition behind using hyperlinks to build the graph is that the more frequently one source links to another (i.e., the higher $w(s, s')$), the higher the chances of a random reader to (click and) reach the reliable/unreliable source $s'$ from $s$. Notably, hyperlinks also serve as a proxy for content-based interactions, as they are typically used to cite content from the referred article. Thus, a higher $w(s, s')$ also implies a stronger content-based relationship. Therefore, this simple weighted, hyperlink-based, and source-centered approach potentially captures both interaction types among news sources simultaneously, while being relatively easy to scale.

\subsection{Reinforcement Learning Strategy}
\label{subsec:rl-algorithms}

Our reinforcement learning reliability framework models reliability in terms of a Markov Decision Process (MDP).
An MDP is defined by a 4-tuple $\langle\mathbb{S}, A, P_a, r_a\rangle$ where $\mathbb{S}$ is a set of states, $A$ a set of actions, $P_a(s, s')$ is the probability that action $a$ in state $s$ will lead to state $s'$, and $r_a(s, s')$ is the immediate reward perceived after taking action $a$ in state $s$ leading to state $s'$~\citep{sutton2018reinforcement,puterman2014markov,kaelbling1996reinforcement}.

Given an MDP, a decision process is then specified by defining a policy $\pi$ that provides the probability of taking action $a$ in state $s$.
In turn, given a policy $\pi$ we can estimate the value of each state, $V^\pi(s)$, in terms of how good it is to be in that state following the policy.
In particular, the value $V^\pi(s)$ is given by the \textit{Bellman equation} which is defined, for any state $s \in \mathbb{S}$, recursively as:
\begin{equation}
\label{eq:value}
    V^\pi(s) = \sum_{s'\in\mathbb{S}}{P^\pi(s, s')[r(s') + \gamma V^\pi(s')]}
\end{equation}
where $\gamma \in [0, 1)$ is known as the \textit{discount factor} and $r(s)$ is the immediate reward received when reaching $s$. Thus, we address the reliability estimation as an MDP $\langle\mathbb{S}, A, P, r\rangle$ such that:
(a) The set of states $\mathbb{S}$ are all the news media websites on the Web ---\textit{i.e.} we have $\mathbb{S} = S$;
(b) The set of actions $A$ contains only one element, the \textit{"move to a different news media website"} action;
(c) The probability $P$ of moving from $s$ to $s'$ will be given by the proportion of hyperlinks in $s$ connecting to $s'$ ---\textit{i.e.} we have $P(s, s') = w(s, s')$; and 
(d) The reward $r$ of moving to a source is determined only by the source itself, and it will be positive or negative for known reliable or unreliable sources respectively ---\textit{i.e.} we have $r(s, s') = r(s')$ where $r:S\mapsto\mathbb{R}$ such that $r(s) = 1$ if $s\in\hat{S^+}$; $r(s) = -1$ if $s\in\hat{S^-}$; $r(s) = 0$ otherwise.
In simple words, we can think of modeling the problem as if there was a ``virtual user'' browsing from one news media source to another with probability proportional to how strongly connected they are, and who will perceive a positive or negative signal (the reward) when arriving to known reliable or unreliable sources, respectively.
Given this framework, now the challenge is how to estimate the \textit{reliability scores} $\rho(s)$.

\subsubsection{Perceived Future Reliability}
\label{subsec:value-iter}

\begin{algorithm}[t]
\caption{\textit{F-Reliability} strategy for $\rho(s)$.}\label{alg:value-iteration}
    \begin{algorithmic}
        \State Set $\forall s\in S, \rho(s)=0$
        \Repeat
            \State $\Delta = 0$
            \ForAll {$s \in S$}
                \State $\rho'(s) = \sum_{s'\in S}{P(s, s')[r(s') + \gamma \rho(s')]}$
                \State $\Delta = max(\Delta, |\rho'(s) - \rho(s)|)$
            \EndFor
            \State $\rho = \rho'$
        \Until{$\Delta$ \textit{is small enough}}
    \end{algorithmic}
\end{algorithm}

Under this simple framework, our initial approach involves estimating reliability by ``looking to the future''.
To be more precise, we will assume the reliability degree $\rho(s)$ is proportional to the \textit{expected} perceived reliability (reward) by the virtual user.
Consequently, a source is considered more reliable (or unreliable) if it is expected to guide the virtual user to well-known reliable (or unreliable) sources.

To achieve this, we can simply set $\rho(s) = V(s)$, as Equation~\ref{eq:value} defines $V(s)$ as the discounted long-term future rewards received following a policy $\pi$.
Note that, given that we only have one possible action in $\mathbb{A}$, the policy $\pi$ is trivial and thus $P^\pi(s, s') = P(s, s')$.
Therefore, a source $s$ will have a higher (lower) $\rho(s)$ the more positive (negative) its total expected future reward $V(s)$.
In other words, it reflects how much $s$ is expected to guide us to known reliable (unreliable) sources, as intended.
We will refer to this strategy as \textit{``F-Reliability''}.
In practice, the computation of $V(s)$ can be done using the \textit{Value Iteration} algorithm~\citep{sutton2018reinforcement}.
Thus, we compute $\rho(s)$ for our specific MDP as shown in Alg. \ref{alg:value-iteration}.

\setlength{\belowdisplayskip}{0pt} \setlength{\belowdisplayshortskip}{0pt}
\setlength{\abovedisplayskip}{0pt} \setlength{\abovedisplayshortskip}{0pt}
\subsubsection{Accumulated Past Reliability}
\label{subsec:reverse-iter}

An alternative approach is to estimate reliability by ``looking to the past'' rather than the future.
Specifically, we assume that the reliability degree $\rho(s)$ is proportional to the accumulated reliability (reward) perceived by the virtual user in reaching the current source $s$.
Consequently, a source becomes more reliable (unreliable) as more known reliable (unreliable) sources lead to it.

To formalize the above intuition, we leverage the \textit{reverse Bellman equation} introduced by \citet{yao2013reinforcement}. 
This equation is recursively defined for any state $s \in \mathbb{S}$ as:
\begin{equation}
\label{eq:reverse}
    R^\pi(s) = r(s) + \gamma\sum_{s'\in\mathbb{S}}{P^\pi(s', s)R^\pi(s')}
\end{equation}

In contrast to Equation~\ref{eq:value} that looks forward from a state to define its value, this equation looks backward to define it ---note $P^\pi(s, s')$ is swapped to $P^\pi(s', s)$.
More precisely, while $V(s)$ defines the value of a state based on the forward accumulated reward, $R(s)$ does so in terms of the historical accumulated reward.
Therefore, by setting $\rho(s) = R(s)$, a source $s$ will have a higher (lower) $\rho(s)$ the more positive (negative) is the accumulated reward $R(s)$ ---\textit{i.e.} the more known reliable (unreliable) sources lead to $s$, as intended.
We will refer to this strategy as \textit{``P-Reliability''}.
In practice, we can again employ \textit{Value Iteration} to compute $\rho(s)$ using $R(s)$ as shown in Algorithm~\ref{alg:reverse-iteration}.

\begin{algorithm}[t]
\caption{\textit{P-Reliability} strategy for $\rho(s)$.}\label{alg:reverse-iteration}
    \begin{algorithmic}
        \State Set $\forall s\in S, \rho(s)=0$
        \Repeat
            \State $\Delta = 0$
            \ForAll {$s \in S$}
                \State $\rho'(s) = r(s) + \gamma\sum_{s'\in S}{P(s', s)\rho(s')}$
                \State $\Delta = max(\Delta, |\rho'(s) - \rho(s)|)$
            \EndFor
            \State $\rho = \rho'$
        \Until{$\Delta$ \textit{is small enough}}
    \end{algorithmic}
\end{algorithm}

\subsubsection{Past and Future Perceived Reliability}
\label{subsec:reliability-iter}

Lastly, we can explore an approach that combines both ``the future and the past''.
Intuitively, we can argue that the transfer of reliability between news media sources and their neighboring sources is asymmetric.
Specifically, the impact on the reliability, $\rho(s)$, of a source $s$ when referencing a reliable source is not equivalent to the effect of a reliable source referencing $s$.\footnote{\textit{e.g.} it is not the same for your reputation as a news media if The New York Times references you as you referencing it.}
Moreover, this asymmetry extends to both reliable and unreliable sources.
That is, a reliable source referencing $s$ carries a different weight than an unreliable one referencing $s$, and vice versa.\footnote{\textit{e.g.} The New York Times referencing you has not the same impact on your reputation as a fake news media referencing you, or as you referencing a fake news media.}
In a broader sense, we can think of a source $s$ increasing its reliability $\rho(s)$ as more reliable sources link to it, while losing reliability as it links to more unreliable sources.

To formalize this asymmetric behavior, we can incorporate both $R(s)$ and $V(s)$ into our reliability model.
More precisely, let $V^-(s)$ be $V(s)$ where only negative rewards $r(s)$ are allowed, and analogously, $R^+(s)$ with only positive rewards, then we can define $\rho(s)$ as:
\begin{equation}
\label{eq:reliability}
    \rho(s) = V^-(s) + R^+(s)
\end{equation}
As a result, a source $s$ will have a higher reliability $\rho(s)$ the more reliable sources link to it (\textit{i.e.} the higher $R^+(s)$), and lower reliability the more it links to unreliable sources (\textit{i.e.} the lower $V^-(s)$).
We will refer to this strategy as \textit{``FP-Reliability''}.
\setlength{\belowdisplayskip}{0pt} \setlength{\belowdisplayshortskip}{0pt}
\setlength{\abovedisplayskip}{0pt} \setlength{\abovedisplayshortskip}{0pt}
\subsection{Reliability Investment Strategy}
\label{subsec:invest}

\begin{algorithm}[t]
\caption{\textit{I-Reliability} strategy for $\rho(s)$.}\label{alg:invest}
    \begin{algorithmic}
        \State Set $\forall s\in S, \rho(s) = r(s)$
        \Repeat
            \ForAll {$s \in S$}\Comment{Investment step}
                \State $totalcredits(s) = \sum_{s'\in S}{w(s', s)\rho(s')}$
            \EndFor
            \ForAll {$s \in S$}\Comment{Credit collection step}
                \State $profit = \sum_{s'\in S}{w(s, s') credits_s(s')}$
                \State $\rho(s) = \rho(s) + profit$
            \EndFor
        \Until{$n$ \textit{times}}
    \end{algorithmic}
\end{algorithm}

A well-established algorithm used in the field of \textit{truth discovery}~\citep{li2016survey} is the \textit{Investment} algorithm~\citep{pasternack-roth-2010-knowing}.
This algorithm is an iterative method in which two interdependent steps are repeated: (1) sources uniformly ``invest'' their trustworthiness among their claimed values; (2) sources collect credits back from the claimed values which update, in turn, their trustworthiness.
Inspired by this ``invest and collect'' intuition, we now formulate an algorithm based on the same principle.
Initially, each source will distribute its reliability $\rho(s)$ among neighboring sources in proportion to the strength of their links, \textit{i.e.} $\propto w(s, s')$.
In essence, during the investment step, the total credits invested in each source $s$ is defined as follows:
\begin{equation}
totalcredits(s) = \sum_{s'\in S}{w(s', s)\cdot\rho(s')}
\end{equation}
In the subsequent credit collection step, the total credits are distributed among investors, $s'$, in proportion to their contribution to the source $s$:
\begin{equation}
credits_{s'}(s) = w_{s'}(s) \cdot totalcredits(s)
\end{equation}
Here, $w_{s'}(s)\in[0, 1]$ represents the proportion of total \textit{inbound} hyperlinks in $s$ originating from $s'$.
The reliability degree is then updated by collecting the credits back in proportion to the invested percentage:
\begin{equation}
\rho(s) = \rho(s) + \sum_{s'\in S}{w(s, s')\cdot credits_{s}(s')}
\end{equation}
Finally, we repeat this process $n$ times to update $\rho(s)$ considering values from up to $n$-hop-away sources in the graph, as illustrated in Algorithm~\ref{alg:invest}.



\section{Data}
\label{subsec:data} 
\subsection{A real-world scale news media graph}
\label{subsec:graph}

\begin{table}
    \centering
    \small
    \begin{tabular}{cc|cc}
        \toprule
        \multicolumn{2}{c}{\textbf{CC-News}} & \multicolumn{2}{c}{\textbf{Graph}} \\
        \textbf{snapshot} & \textit{\#articles} & \textit{\#nodes} & \textit{\#edges} \\
        \midrule
        2019/08 & 17M & 6,799 & 171,810 \\
        2020/08 & 23M & 11,427 & 276,666 \\
        2021/08 & 28M & 10,938 & 315,447 \\
        2022/08 & 35M & 10,607 & 354,386 \\
        \midrule
        \textit{all above} & 103M & 17,057 & 909,354 \\
        \bottomrule
    \end{tabular}
    \caption{CC-News snapshots and the obtained graphs. The last row corresponds to our final graph.}
    \label{tab:graphs}
\end{table}

The \textit{Common Crawl Foundation}\footnote{\url{https://commoncrawl.org}} maintains the \textit{Common Crawl News Dataset} (\textit{CC-News}), the world's largest collection of news articles crawled from global news web sites since 2016.
The data is updated daily and published as a series of snapshots organized by year and month.

We developed a Python CC-News processing pipeline that takes care of building the news media graph, $G$, from CC-News snapshots (details in Appendix~\ref{app:graph-pipeline}).
Similar to the \textit{CCNet} pipeline~\cite{wenzek-etal-2020-ccnet}, our pipeline utilizes the language classifier from fastText~\citep{joulin-etal-2017-bag, grave-etal-2018-learning} to categorize news articles into 176 languages. 
Consequently, for a given CC-NEWS snapshot URL, the pipeline generates one graph for each supported language showing how news sources relate to each other in that language.
However, in this paper, we focus exclusively on the English graph due to the predominance of available ground truth data for experimentation in this language.
Specifically, for experimentation, we will use the English graph obtained from joining four different \textit{CC-News} snapshots corresponding to August over the past four years (2019 to 2022).\footnote{By selecting the same month, we ensure a consistent 4-year time span while limiting the processed news articles to approximately 100M.}
As indicated in Table~\ref{tab:graphs}, this process resulted in a unified graph containing around 17k English-speaking news media sources and nearly 1M connections ---graph shown in Figure~\ref{fig:graph} (Appendix \ref{app:ablation}).

\subsection{Ground truth datasets}
\label{subsec:datasets}

To facilitate a comparative analysis with previous studies on the source reliability classification task, we employ the largest dataset published earlier by \citet{baly-etal-2018-predicting}.
This dataset encompasses 1066 annotated news media URL domains extracted from \textit{Media Bias/Fact Check} (MBFC) ---refer to the first row of Table~\ref{tab:datasets} for details.\footnote{Original factuality labels were transformed into reliability labels following  \citet{norregaard2020nela} strategy.}
Furthermore, for a more comprehensive evaluation, we employ an extended dataset meticulously created by merging ground truth labels from various sources, as outlined below:


\noindent
\textbf{$\bullet$ MBFC:} 
we followed a similar process as in \citet{baly-etal-2018-predicting} but crawling the entire MBFC website to extract 4138 ground truth labels.
Following \citet{norregaard2020nela}, we aggregated these labels into three classes: ``reliable'' for sources with high or very high factual reporting, ``unreliable'' for sources flagged as conspiracy, pseudoscience, or with low/very low factual reporting, and ``mixed'' for sources with mixed factual reporting.


\noindent
\textbf{$\bullet$ Wikipedia's perennial sources:} the platform hosts a list of sources discussed by the community regarding their reliability and use on the platform.\footnote{\url{https://en.wikipedia.org/wiki/Wikipedia:Reliable_sources/Perennial_sources}}
We extracted 553 ground truth labels from this list applying the following policy:
sources marked as \textit{generally reliable} were labeled as ``reliable'';
sources marked as \textit{generally unreliable}, \textit{deprecated}, or \textit{blacklisted} were labeled as ``unreliable''; and sources marked as \textit{no consensus}, \textit{stale discussions} or \textit{discussion in progress} as ``mixed'';


\noindent
\textbf{$\bullet$ Fake news:} we manually collected a list of 556 unreliable sources from fake news websites, including the Wikipedia list of fake news websites\footnote{\url{https://en.wikipedia.org/wiki/List_of_fake_news_websites}} and a report from the Institute for Strategic Dialogue identifying active and inactive fake news domains~\citep{fakenews2020}.


\noindent
\textbf{$\bullet$ NewsGuard:} a paid rating service similar to MBFC, provides both a verdict and a reliability score based on predefined journalistic criteria (details in Appendix~\ref{app:newsguard}).
Due to license limitations, we could only use the 85 ground truth values included in the NELA-GT-2018 dataset~\citep{norregaard2019nela}.
However, as detailed in Section~\ref{subsec:correlation}, the inclusion of NewsGuard enables us to measure the correlation between the estimated reliability degrees $\rho(s)$ and the scores provided by journalists.


\begin{table}
    \centering
    \small
    \begin{tabular}{c@{}ccc}
        \toprule
        & \multicolumn{3}{c}{\textbf{Label distribution}} \\
        \textbf{Dataset} & \textit{unreliable} & \textit{mixed} & \textit{reliable} \\
        \midrule
         \citet{baly-etal-2018-predicting} & 256 & 268 & 542 \\
         Our own & 1425 & 1461 & 2446 \\

         \addlinespace
         \cdashline{1-4}
         \addlinespace

         \textit{MBFC} & \textit{546} & \textit{1363} & \textit{2229} \\
         \textit{Wikipedia} & \textit{298} & \textit{98} & \textit{157} \\
         \textit{Fake News} & \textit{556} & - & - \\
         \textit{NewsGuard} & \textit{25} & - & \textit{60} \\
        \bottomrule
    \end{tabular}
    \caption{Datasets details. Bottom part shows individual contributions to our final dataset.}
    \label{tab:datasets}
\end{table}


Hence, our final aggregated dataset comprises 5332 news URL domains, each annotated with 3-class reliability labels.
As illustrated in Table~\ref{tab:datasets}, its scale surpasses that of the largest one to date~\citep{baly-etal-2018-predicting}, being an order of magnitude larger. 
For evaluation, it is crucial that the source $s$ is present in the graph, as we want to assess how well the reliability degree $\rho(s)$ is computed from it.
Therefore, we limit our experimentation to using the subset of the ground truth dataset corresponding to the nodes within our graph.
This subset contains approximately $40\%$ of the total ground truth sources.
In particular, 400 sources from \citet{baly-etal-2018-predicting} (294 "reliable," 85 "mixed," and 21 "unreliable") and 2117 sources from our own dataset (1630 "reliable," 321 "mixed," and 166 "unreliable").
Additionally, since our goal is to evaluate the ability of $\rho(s)$ to distinguish reliable from unreliable sources (see conditions 1 and 2 in Section~\ref{subsec:problem-formulation}), we merge "unreliable" and "mixed" labels to create the following three experimentation sets:

\noindent
\textbf{$\bullet$ ExpsetA:} 294 \textit{reliable} and 106 \textit{unreliable} sources from \citet{baly-etal-2018-predicting}.

\noindent
\textbf{$\bullet$ ExpsetB:} 1630 \textit{reliable} and 487 \textit{unreliable} sources from our dataset.

\noindent
\textbf{$\bullet$ ExpsetB$^-$:} 1630 \textit{reliable} and 166 \textit{unreliable} sources. A simpler version of \textit{ExpsetB} removing ``mixed'' from \textit{unreliable} sources.

\section{Experiments and Evaluation Results}
\label{sec:experimentation}
For experimentation, we define the reward values based on ground truth labels as $r(s) = 1$ if the label is ``reliable'', $r(s) = -1$ if ``unreliable'', and $r(s) = 0$ otherwise.
In addition, selecting appropriate hyperparameter values is crucial.
For \textit{I-Reliability}, $n$ controls how far to look in the neighborhood for investments (how many nodes away).
Similarly, in the reinforcement learning strategies, the discount factor $\gamma$ controls the distance of looking back/forward; $\gamma\approx0$ focuses mostly on present reward $r(s)$, while $\gamma\approx1$ considers all history/future to compute $\rho(s)$.
We performed a grid search over $n \in [1, 10]$ and $\gamma\in [0.05, 0.95]$ to determine the best hyperparameter values on each of the three experimental sets.
The grid search was performed using 5-fold cross validation selecting the $n$ and $\gamma$ that obtained the best \textit{macro avg. F$_1$} on the reliability classification task, as described in Section \ref{subsec:classification}.
We observed that, independently of the dataset, \textit{better reliability estimation is achieved when looking mostly at nearby sources}, as better performance was obtained with small $n$ ($n \leq 2$) and $\gamma$ ($\gamma < 0.5$) values ---details in Appendix~\ref{app:hyperparameters}.



\subsection{Reliability Classification Results}
\label{subsec:classification}

\begin{table}[t]
    \centering
    \small
    \begin{tabular}{c@{~~}c@{~~~}c@{~~~}c@{~~~}c@{~~~}c}
        \toprule
        &\multirow{2}{*}{\textbf{Strategy}} & \multicolumn{3}{c}{\textbf{F$_1$ score}} &  \\
        \cmidrule(lr){3-5}
        \textbf{Data} &  & \textit{macro avg.} & \textit{reliable} & \textit{unreliable} & \textbf{Acc.} \\
        \midrule
        \multirow{8}{*}{\rotatebox{90}{\textbf{ExpsetA}}} & M-BL & 42.33 & 84.66 & 0.00 & 73.44 \\
        & R-BL & 48.85 & 61.76 & 35.94 & 52.33 \\
        & Baly18 & 67.87 & 84.81 & 50.92 & 76.95 \\
        & Baly20 & 65.24 & 82.99 & 47.50 & 74.37 \\
        \cmidrule(lr){2-6}
        & \textit{F-R} & 61.52 & 87.62 & 35.42 & 79.26 \\
        & \textit{P-R} & \textbf{72.67} & \underline{\textbf{90.05}} & \textbf{55.29} & \underline{\textbf{83.79}} \\
        & \textit{FP-R} & 69.28 & 89.23 & 49.34 & 82.29 \\
        & \textit{I-R} & \underline{\textbf{72.81}} & \textbf{90.03} & \underline{\textbf{55.60}} & \textbf{83.77} \\

        \midrule

        \multirow{6}{*}{\rotatebox{90}{\textbf{ExpsetB}}} & M-BL & 43.50 & 87.00 & 0.00 & 77.00 \\
        & R-BL & 47.48 & 62.17 & 32.80 & 51.63 \\
        \cmidrule(lr){2-6}
        & \textit{F-R} & 61.85 & 79.72 & 43.98 & 70.34 \\
        & \textit{P-R} & \textbf{74.69} & \textbf{88.29} & \textbf{61.10} & \textbf{82.00} \\
        & \textit{FP-R} & 55.95 & 68.08 & 43.82 & 59.38 \\
        & \textit{I-R} & \underline{\textbf{75.51}} & \underline{\textbf{89.30}} & \underline{\textbf{61.72}} & \underline{\textbf{83.28}} \\

        \midrule

        \multirow{6}{*}{\rotatebox{90}{\textbf{ExpsetB$^-$}}} & M-BL & 47.58 & 95.15 & 0.00 & 90.76 \\
        & R-BL & 39.17 & 63.04 & 15.31 & 48.55 \\
        \cmidrule{2-6}
        & \textit{F-R} & 62.18 & 90.23 & 34.12 & 83.02 \\
        & \textit{P-R} & \textbf{78.90} & \textbf{95.83} & \textbf{61.97} & \textbf{92.48} \\
        & \textit{FP-R} & 59.20 & 84.66 & 33.74 & 75.11 \\
        & \textit{I-R} & \underline{\textbf{81.05}} & \underline{\textbf{96.71}} & \underline{\textbf{65.39}} & \underline{\textbf{93.99}} \\

        \bottomrule
    \end{tabular}
    \caption{5-fold cross-validation average results for reliability classification.
    The best-performing values are \underline{\textbf{underlined}}, while the 2nd-best results appear in \textbf{bold} font.
    R-BL and M-BL refer to random and majority class baselines; Baly18 and Baly20 refer to \citet{baly-etal-2018-predicting} and \citet{baly-etal-2020-written}; and \textit{*}-R stands for \textit{*}-Reliability.
    }
    \label{tab:results}
\end{table}

In this section, we focus on evaluating the first two conditions for $\rho(s)$ given in Section~\ref{subsec:problem-formulation}.
These conditions allow us to measure the ability of $\rho(s)$ to distinguish reliable from unreliable sources.
For comparison, we follow the evaluation procedure from \citet{baly-etal-2018-predicting} and report results for 5-fold cross-validation.
More precisely, in each k-fold iteration, we only use ground truth rewards $r(s)$ from four folds to compute $\rho(s)$ for all 17k sources in the graph, and using conditions 1 and 2, all $s$ in the hold-out fold are classified as \textit{reliable} ($\rho(s) > 0$) or \textit{unreliable} ($\rho(s) \leq 0$). 

Table~\ref{tab:results} shows the evaluation results obtained on the three experimentation sets along with two naive baselines for reference, random and majority class classifiers.\footnote{For a comprehensive view of additional metrics, such as precision, recall, and confidence intervals, refer to Table~\ref{tab:results-full} in the Appendix.}
In addition, for \textit{ExpsetA}, we also report the results obtained using the classification models introduced in previous works~\citep{baly-etal-2018-predicting, baly-etal-2020-written}.
These classifiers combine multiple content-based, audience-based, and metadata-based features about the sources.
Authors released the pre-computed features values for the \citet{baly-etal-2018-predicting} dataset and thus, using their source code,\footnote{\href{https://github.com/ramybaly/News-Media-Reliability}{\url{github.com/ramybaly/News-Media-Reliability}}.} we were able to train and evaluate their classifiers on the \textit{ExpsetA} set.
However, since building these features relies on multiple external sources (\textit{e.g.} Twitter, Facebook, YouTube, etc.), we were not able to evaluate their method in our new dataset given current API restrictions to access them.

Observing the performance across the different datasets, all four strategies outperform the random and majority class baselines by a statistically significant difference (paired $t$-test with $p\text{\textit{-value}}<0.02$). 
In addition, both \textit{P-Reliability} and \textit{I-Reliability} consistently outperform other strategies, including those presented in previously published works ($p\text{\textit{-value}}<0.03$), however, the difference between these two strategies is not statistically significant ($p\text{\textit{-value}}> 0.5$).
From the results we can also see that, regardless of the chosen strategy and dataset, the $F_1$ score for the \textit{unreliable} class consistently remains lower when compared to the \textit{reliable} class.
This suggests that identifying unreliable sources is more challenging, likely due to the dataset imbalance favoring reliable sources ---note that in both \textit{ExpsetA} and \textit{ExpsetB}, only approximately $25\%$ of the sources are unreliable. This imbalance results in models having fewer negative signals (\textit{i.e.} rewards $r(s) = -1$) to learn to identify unreliable sources effectively.
Another contributing factor is the inclusion of ``mixed'' labels in the \textit{unreliable} group, making the task more challenging by incorporating unreliable sources whose reliability is not clearly defined.
This hypothesis is supported by the results from \textit{ExpsetB$^-$}, where the removal of ``mixed'' labels results in improvements across all metrics ---note that the highest \textit{unreliable} $F_1$ score is achieved while the dataset is significantly more unbalanced (ten times fewer unreliable sources than reliable ones).
Concerning the reinforcement learning strategies, \textit{P-Reliability} significantly ($p\text{\textit{-value}}\le 0.02$) outperformed \textit{F-Reliability} suggesting that the reliability of a source \textit{is more significantly influenced by its origins than by the destinations it reaches}.
On the other hand, \textit{FP-Reliability} shows poor performance, mainly due to the different nature of $V^-(s)$ and $R^+(s)$ in Equation~\ref{eq:reliability} ---note that $V(s)$ is defined as an expectation, whereas $R(s)$ is not.\footnote{Transition probabilities are not normalized in both directions, only in the forward direction. Consequently, there is no inherent mathematical symmetry that ensures $V^-(s)$ and $R^+(s)$ will equilibrate to zero ($\rho(s) = 0$) when they correspond to an equivalent number of unreliable and reliable sources}

\begin{table}[t]
    \centering
    \small
    \begin{tabular}{r@{~~~}c@{~~~}c@{~~~}c@{~~~}c}
        \toprule
        \multirow{2}{*}{\textbf{Strategy}} & \multicolumn{3}{c}{\textbf{F$_1$ score}} &  \\
        \cmidrule(lr){2-4}
        & \textit{macro avg.} & \textit{reliable} & \textit{unreliable} & \textbf{Acc.} \\
        \midrule
        \multicolumn{1}{l}{\textit{P-Reliability}} & \textit{72.67} & \underline{\textbf{90.05}} & \textit{55.29} & \textit{\underline{\textbf{83.79}}} \\
        \cmidrule(lr){2-5}
        \textit{+Baly18} & \textbf{77.11} & 87.75 & \textbf{66.47} & 82.11 \\
        \textit{+Baly20} & 74.36 & 86.02 & 62.70 & 79.69 \\
        \midrule
        \multicolumn{1}{l}{\textit{I-Reliability}} & \textit{72.81} & \textbf{90.03} & \textit{55.60} & \textit{\textbf{83.77}} \\
        \cmidrule(lr){2-5}
        \textit{+Baly18} & \underline{\textbf{77.47}} & 87.89 & \underline{\textbf{67.06}} & 82.34 \\
        \textit{+Baly20} & 72.88 & 85.46 & 60.30 & 78.74 \\
        \bottomrule
    \end{tabular}
    \caption{Ensemble results for \textit{P-Reliability} and \textit{I-Reliability} strategies on \textit{ExpsetA}. The best performance results are \underline{\textbf{underlined}}, while the 2nd-best appear in \textbf{bold} font.}
    \label{tab:results-ensemble}
\end{table}

Finally, to assess the complementarity of our graph-based strategies with content-based ones, we performed an additional experiment: a simple voting ensemble between our strategies and Baly's models. Specifically, sources were classified as reliable only when both models agreed on the classification. Table~\ref{tab:results-ensemble} presents the obtained results for our two best-performing models, \textit{P-Reliability} and \textit{I-Reliability}, on \textit{ExpsetA}.\footnote{Full results included in Table~\ref{tab:results-full} (Appendix).} The ensemble approach enabled the models to further improve their performance, particularly in detecting unreliable sources, achieving the highest \textit{macro avg.} $F_1$ score on this dataset (77.47). 

\subsection{Correlation with human judgment}
\label{subsec:correlation}

In this section, we focus on evaluating the final condition in the definition of $\rho(s)$ in Section~\ref{subsec:problem-formulation}.
This condition enables $\rho(s)$ to assess the reliability of $s$ \textit{relative} to other sources, allowing the ranking of sources based on their reliability degrees.
For evaluation, we use the \textit{NewsGuard} dataset introduced in Section~\ref{subsec:datasets} containing the 85 ground truth reliability scores provided by trained journalists.
The score ranges from 0 to 100 and is obtained by answering 9 questions that address different journalistic criteria ---details in Appendix~\ref{app:newsguard}.
Table~\ref{tab:ranking} shows examples of NewsGuard scores and their estimated reliability degree.\footnote{For ease of comparison, in this table, $\rho(s)$ is normalized in the range $[-1, 1]$ by dividing it by the maximum (when $\rho(s) \geq 0$) and minimum value (when $\rho(s) < 0$).}
 
\begin{table}[t]
    \centering
    \small
    \begin{tabular}{c@{}cc|c}
        \toprule
        \textbf{Rank} & \textbf{Domain} & \textbf{Score} & $\hat{\rho}(s)$ \\
        \midrule
        1 & \href{https://bbc.co.uk}{\url{bbc.co.uk}} & 100.0 & 0.995 \\
        2 & \href{https://cnbc.com}{\url{cnbc.com}} & 95.0 & 0.995 \\
        3 & \href{https://dailysignal.com}{\url{dailysignal.com}} & 92.5 & 0.830 \\
        4 & \href{https://thinkprogress.org}{\url{thinkprogress.org}} & 90.0 & 0.907 \\
        5 & \href{https://independent.co.uk}{\url{independent.co.uk}} & 87.5 & 0.968 \\
        \midrule
        1 & \href{https://sputniknews.com}{\url{sputniknews.com}} & 7.5 & -0.992 \\
        2 & \href{https://truepundit.com}{\url{truepundit.com}} & 12.5 & -0.957 \\
        3 & \href{https://dailymail.co.uk}{\url{dailymail.co.uk}} & 15.0 & -0.998 \\
        4 & \href{https://theduran.com}{\url{theduran.com}} & 17.5 & -0.954 \\
        5 & \href{https://thegatewaypundit.com}{\url{thegatewaypundit.com}} & 20.0 & -0.994 \\
        \bottomrule
    \end{tabular}
    \caption{NewsGuard top-5 unique most scored (top part) and least scored sources (bottom part) along with the estimated $\rho(s)$ given by \textit{P-Reliability}.}
    \label{tab:ranking}
\end{table}

We measure the correlation between these human-provided scores and their estimated reliability degree by computing the \textit{Pearson correlation coefficient} (PCC) and the \textit{Spearman's rank correlation coefficient} (SRCC).
PCC measures the linear relationship between two variables, whereas SRCC assesses the monotonic relationship, making it \textit{more suitable for evaluating the relative reliability} of sources, as it captures ranked associations regardless of the exact numerical values (condition 3 for $\rho$).
In particular, we perform the evaluation under two scenarios, when sources are known to be (un)reliable and, more challenging, when their reliability is not known in advance.\footnote{For instance, is $\rho(\text{\textit{cnbc.com}}) < \rho(\text{\textit{bbc.co.uk}})$? that is, is ``bbc.co.uk'' more reliable than the ``cnbc.com''? knowing in advance that both are reliable ($r(\text{\textit{cnbc.com}})=r(\text{\textit{bbc.co.uk}})=1$) vs. not knowing it ($r(\text{\textit{cnbc.com}})=r(\text{\textit{bbc.co.uk}})=0$).}
In other words, we perform the evaluation following, respectively, two experimental settings: ($\clubsuit$) we use all the ground truth rewards from the largest experimental set, \textit{ExpsetB}, to learn the reliability degree $\rho(s)$ of all 17k sources in the graph; and ($\diamondsuit$) we repeat the same process but removing the rewards for all the 85 domains used for evaluation.


\begin{table}[t]
    \centering
    \small
    \begin{tabular}{c|c@{~~}c|c@{~~}c}
        \toprule
        \textbf{Strategy} & \textbf{PCC} & \textbf{\textit{p-value}} & \textbf{SRCC} & \textbf{\textit{p-value}} \\
        \midrule
        Random baseline & 0.058 & \textit{0.6} & 0.066 & \textit{0.6} \\
        PageRank baseline & 0.313 & \textit{0.008} & 0.544 & \textit{1e-06} \\
        \midrule
         \textit{F-Reliability}$^\diamondsuit$ & 0.556 & \textit{5e-07} & 0.295 & \textit{1e-02} \\
         \textit{P-Reliability}$^\diamondsuit$ & \textbf{0.647} & \textit{1e-09} & 0.668 & \textit{2e-10} \\
         \textit{FP-Reliability}$^\diamondsuit$ & 0.636 & \textit{3e-09} & \textbf{0.677} & \textit{3e-09} \\
         \textit{I-Reliability}$^\diamondsuit$ & 0.589 & \textit{7e-08} & 0.657 & \textit{5e-10} \\

         \addlinespace
         \cdashline{1-5}
         \addlinespace

         \textit{F-Reliability}$^\clubsuit$ & 0.927 & \textit{1e-30} & 0.544 & \textit{1e-06} \\
         \textit{P-Reliability}$^\clubsuit$ & 0.912 & \textit{9e-33} & \underline{\textbf{0.801}} & \textit{6e-17} \\
         \textit{FP-Reliability}$^\clubsuit$ & \underline{\textbf{0.929}} & \textit{8e-32} & 0.775 & \textit{7e-15} \\
         \textit{I-Reliability}$^\clubsuit$ & 0.757 & \textit{2e-19} & 0.792 & \textit{8e-12} \\
        \bottomrule
    \end{tabular}
    \caption{Correlation between $\rho(s)$ and journalist-provided reliability scores. $\clubsuit$: w/ rewards; $\diamondsuit$: w/o rewards.}
    \label{tab:results-correlation}
\end{table}

Table~\ref{tab:results-correlation} shows the obtained results along with two baselines for reference, random and \textit{PageRank} algorithm~\citep{brin1998107},\footnote{Note that PageRank is unsuitable for classification experiments as its non-negative scores always predicted the positive class, reducing it to a majority-class classifier. Hence, its exclusion from Table~\ref{tab:results}.} scatter plots in Appendix~\ref{app:results}.
We can observe that, as expected, correlations are weaker under the hardest scenario without rewards, specially in terms of PCC which is more sensitive to the bias introduced by the rewards.\footnote{Note that sources with $r(s)=1$ will naturally tend to have a final $\rho(s)$ close to $1$ while sources with $r(s)=-1$ close to $-1$, this bias is heavily reduced when instead of using the actual $\rho(s)$ value we use its ranking (as in SRCC).}
Nevertheless, in both scenarios, obtained correlation coefficients are statistically significant ($p\text{\textit{-value}}\le5\text{e-}07$ for PCC, $p\text{\textit{-value}}\le1\text{e-}06$ for SRCC) and higher than the baselines, except for \textit{F-Reliability}.
In general, the strategies that correlate more strongly with the journalist-provided scores are \textit{P-Reliability} and \textit{FP-Reliability}, showing both a strong linear (PCC) and ranking-based (SRCC) relationship independently of whether rewards were used or not.\footnote{In fact, we performed an additional experiment in which we set $\rho(s) = \frac{\rho_p(s)+\rho_{fp}(s)}{2}$, i.e., the reliability degree was defined as the average of the values obtained by the best performing strategies, \textit{P-Reliability} and \textit{FP-Reliability}, obtaining the strongest correlation values (PCC=0.933, SRCC=0.803 and PCC=0.715, SRCC=0.697 with and without rewards, respectively).}
\textit{FP-Reliability} results suggest that combining \textit{F-Reliability} and \textit{P-Reliability} strategies could be advantageous for estimating \textit{relative} reliability.\footnote{In contrast to reliability classification, where $\rho(s) = 0$ is the fixed threshold separating reliable from unreliable sources.}
Overall, results are inspiring considering that the learning process for all $\rho(s)$ values in the graph leverages only a subset of binary ground truth rewards ($r(s)=-1$ or $1$), \emph{without any explicit notion of ground truth score or degree}. In contrast to reliability scores derived from various qualitative journalistic criteria, $\rho(s)$ approximates the reliability degree solely based on the propagation of these initial rewards throughout the network's structure.\footnote{We are releasing the list of estimations for all 17k sources along with this paper.}

\section{Conclusion and Future Work}
\label{sec:conslusion}

In this study, we introduced an approach for assessing the reliability of news media through their network interactions.
This approach diverges from previous models that depend on content, audience feedback, and/or metadata.
Moreover, unlike in previous works, our method estimates a \textit{reliability degree} rather than a reliability label.
We assessed the quality of the estimated values in terms of reliability classification and correlation with journalists-provided scores.
We found that a source's origins is more indicative of its reliability than its reach and show evidence that it is feasible to predict the reliability of news media using only their network interactions, providing an easier-to-scale approach than prior methods.
As future work, we plan to expand the study by building a larger graph and designing more sophisticated strategies that leverage content-based features. Additionally, we aim to explore the estimation of other news source properties, such as political bias, using the same approach. Finally, we intend to investigate the use of the estimated reliability values in downstream tasks like fact-checking and fake news detection.

\section{Ethical Considerations}
The work presented in this paper has been done in the scope of the CRiTERIA project\footnote{\url{https://www.project-criteria.eu/}} 
that follows the H2020 ethical standards and guidelines.
The Consortium Agreement includes the partners’ commitment to FAIR (findable, accessible, interoperable and re-usable) data management practices and responsible research practices. The framework of the research questions and preliminary results were reviewed by the Project Ethics Check and Audit committee in the form of an on-going work deliverable.

In the present research paper there is no gender bias to be investigated or addressed. The data comes from hyperlinks and URL domain names that can not be associated to a gender. As described in Section \ref{subsec:data}, there is no intentional collection of personal data in any form, and as a consequence, there is no need for data anonymisation or pseudonymisation. Similarly, there is no need of informed and singed consents since there is no direct human participation in the construction of the graphs to calculate the reliability values.

Regarding data and processing security, a snapshot of CC-News data is transferred, held in the local servers and processed for the reliability estimation. The data can be deleted at any time and easily downloaded from the original public CC-News sources (as described in Section \ref{subsec:data}). Any further data processing carried out, is going to be publicly available, along with the reference to this paper. 

Regarding other sources, only data collected by other sources (under Apache or open source licenses) with well described data collection methodology, validated with published results and that is publicly available was used. The overview of the datasets is described in Section \ref{subsec:datasets}, the description includes the annotations distribution in Table \ref{tab:datasets}, and the original sources are properly acknowledge along the paper.

The estimated reliability values were obtained based on initial ground truth labels. This labels capture mainly the factuality of reporting and do not consider other aspects like, for instance, political bias or press freedom rating.
Therefore, computed reliability values should not be considered as \textit{de facto} values.

From a societal perspective, this paper brings a positive impact, improving the situational awareness of decision makers, including fact-checkers. The mathematically defined algorithms are robust to content-related biases since they are both language and content independent (political, religious, racial, etc.).

%

\section{Limitations}


\subsection{Methodology}

The main constraint of the proposed methodology lies in the requirement for news media sources to be included in the graph for their reliability to be calculated. This limitation may arise due to temporal or size constraints in the data used to construct the graph. For instance, a recently emerged news source might not be referenced by others until some time has passed. To address this limitation to a certain degree, assigning $\rho(s) = 0$ to such sources can be considered. This implies that their reliability is indeterminate, indicating an unknown or undetermined status, meaning they are neither reliable nor unreliable.


\subsection{Experimentation}

The main three limitations of the present work regarding the experimentation and evaluation of the proposed approach are:
\begin{enumerate}
    \item \textbf{Only English-speaking news sources:} the proposed methodology is content- and language-independent. However, we focused exclusively on the English-speaking news sources due to the predominance of available ground truth data for experimentation in this language. Further studies need to be done with ground truth for non-English-speaking sources to assess the robustness of the methodology across languages.
    \item \textbf{Restricted graph size:} the graph we used was built processing around 100M news articles from 4 months spanning a 4-year time window.
    This imposes not only a temporal limitation but also restricts the number of sources in the graph.
    However, along with this paper, we release and open source, under Apache 2.0 license, the Python CC-News processing pipeline and the dataset for the community to reproduce the proposed methodology in larger scale.
    \item \textbf{Corpus used to build the graph:} although CC-News is continuously growing on a daily basis, the crawling of news articles started in 2016.
    Therefore, CC-News does not contain articles prior to 2016 and news sources that existed before 2016 but not after, will not be reachable.
    Consequently, for these sources to be included, their articles need to be crawled from a different corpus or manually from the Web.
\end{enumerate}






\section{Acknowledgments}
This work was supported by CRiTERIA, EU project funded under the Horizon 2020 program, grant agreement number 101021866.
We would also like to express our sincere gratitude to the anonymous reviewers for their valuable feedback, which has helped us enhance the quality of this work.




\bibliography{acl_latex}

\newpage
\appendix

\section{Hyperparameter Optimization}
\label{app:hyperparameters}

\begin{figure*}[t]
    \centering
    \includegraphics[width=.9\linewidth]{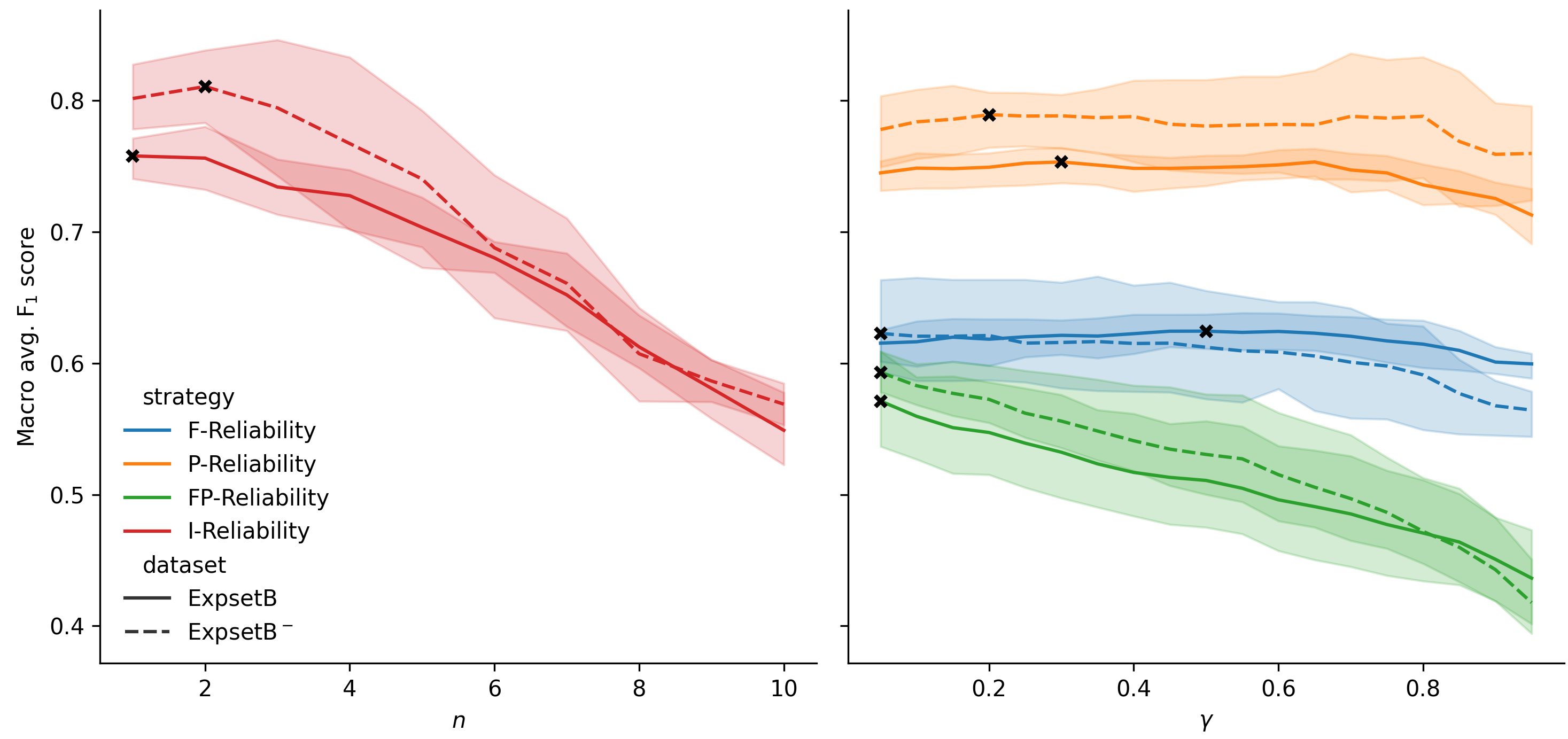}
    \caption{Performance variation across searched values of $n$ (left side) and $\gamma$ (right side) on the \textit{ExpsetB} (solid line) and \textit{ExpsetB}$^-$ (dashed line) datasets. The lines represent the mean values across the 5 folds, and $95\%$ confidence intervals are depicted. Markers highlight selected hyperparameter values.}
    \label{fig:hyperparameters}
\end{figure*}

We performed a grid search to determine the best hyperparameter values on each of the three experimental sets.
More precisely, as in Section \ref{subsec:classification}, the evaluation was performed using 5-fold cross validation on the reliability classification task.
For reinforcement learning strategies, we evaluated $\gamma$ from $0.05$ to $0.95$ in increments of $0.05$ (\textit{i.e.,} $\gamma\in \{0.05,0.1,0.15,\ldots,0.95\}$).
For \textit{I-Reliability}, $n$  from $1$ to $10$ (\textit{i.e.,} $n\in \{1,2,3,\ldots,10\}$).
Finally, the hyperparameter values obtaining the best \textit{macro avg. F$_1$ score} were the one selected on each experimental set. Namely, the selected values for each strategy were:

\begin{itemize}
    \item \textbf{F-Reliability:} $\gamma=0.05$ for \textit{ExpsetA} and \textit{ExpsetB}$^-$, $\gamma=0.5$ for \textit{ExpsetB}.
    \item \textbf{P-Reliability:} $\gamma=0.15$ for \textit{ExpsetA}, $\gamma=0.3$ for \textit{ExpsetB}, and $\gamma=0.2$ for \textit{ExpsetB}$^-$.
    \item \textbf{FP-Reliability:} $\gamma=0.1$ for \textit{ExpsetA}, $\gamma=0.05$ for \textit{ExpsetB} and  \textit{ExpsetB}$^-$.
    \item \textbf{I-Reliability:} $n=1$ for \textit{ExpsetA} and \textit{ExpsetB}, and $n=2$ for \textit{ExpsetB}$^-$.
\end{itemize}

As shown in Figure \ref{fig:hyperparameters}, we can observe that \textit{better reliability estimation is achieved when looking mostly at nearby sources}, as better performance is obtained with small $n$ and $\gamma$ values, namely $n \leq 2$ and $\gamma < 0.5$.
Furthermore, \textit{P-Reliability} (orange line) outperforms the other reinforcement learning strategies, consistently, while being more robust to the choice of $\gamma$, except when $\gamma > 0.7$ from which performance starts to decrease.

Finally, for the \citet{baly-etal-2018-predicting} and \citet{baly-etal-2020-written} classifiers in Table \ref{tab:results}, we follow the same process described by the authors to tune the
SVM hyperparameters, i.e., the cost $C$, the kernel type, and the kernel width $\gamma$ using the 5-fold cross validation maximizing the \textit{F$_1$ score} as with our methods.\footnote{We used the author's source code containing the hyperparameter search in it (\href{https://github.com/ramybaly/News-Media-Reliability}{\url{github.com/ramybaly/News-Media-Reliability}}).}

\section{Temporal Ablation Analysis}
\label{app:ablation}

\begin{figure}[t]
    \centering
    \includegraphics[width=.9\linewidth]{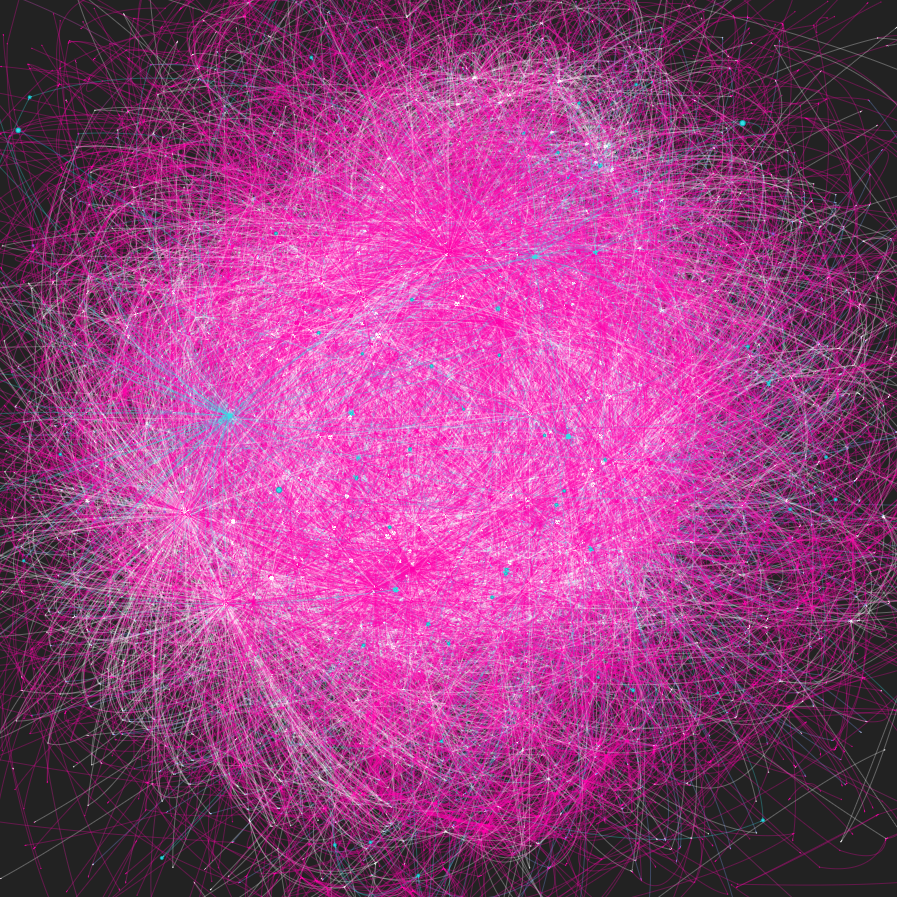}
    \caption{News media graph built from all four CC-News snapshot  (only English-speaking sources) and used for experimentation.}
    \label{fig:graph}
\end{figure}

\begin{figure}[t]
    \centering
    \includegraphics[width=.95\linewidth]{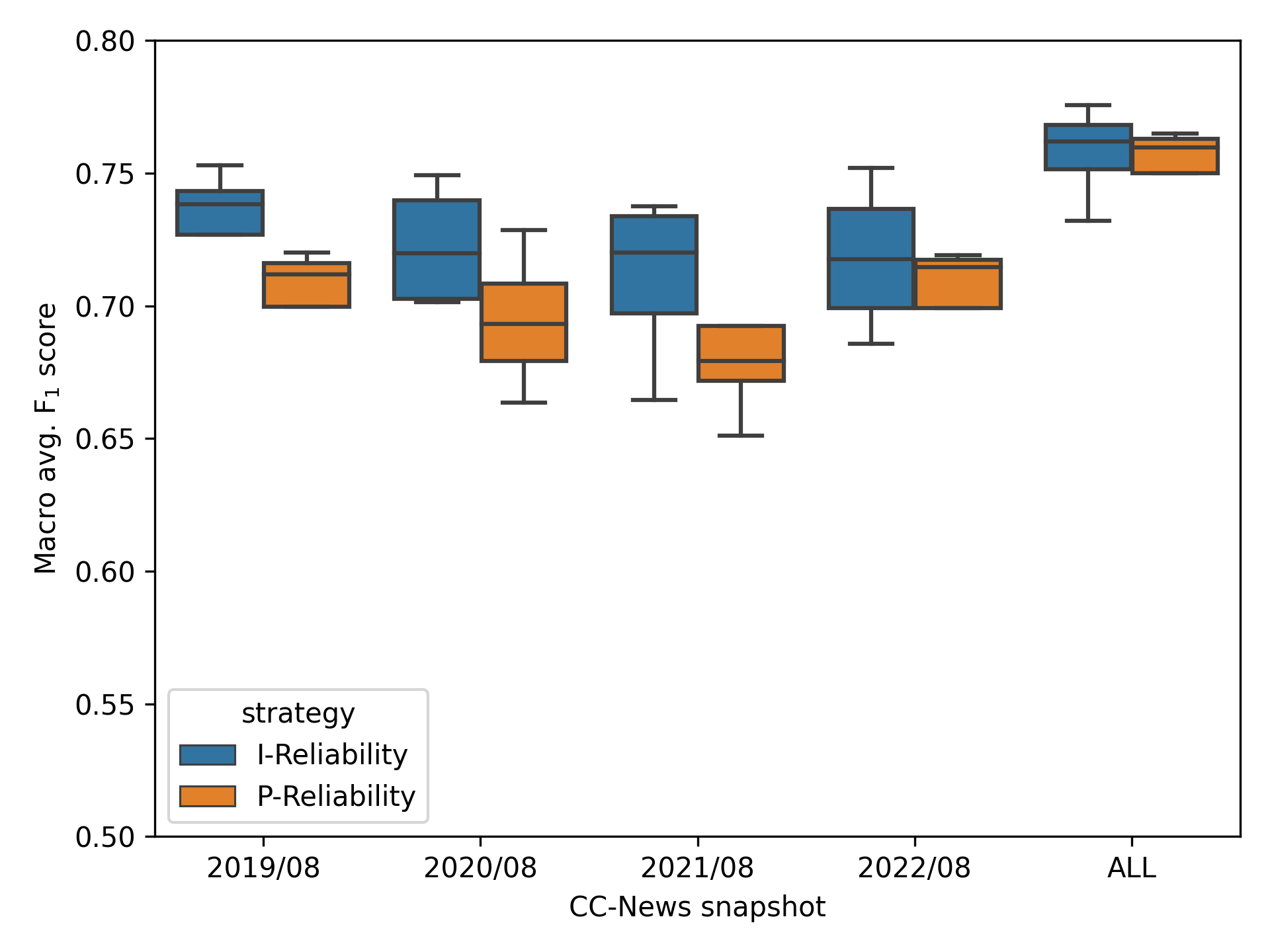}
    \caption{5-fold cross-validation results obtained on the \textit{ExpsetB} dataset with the two best strategies, \textit{P-Reliability} and \textit{I-Reliability}, using different graphs.
    The x-axis represents the CC-News snapshot used to build the graph, and the y-axis the \textit{macro averaged F$_1$ score}. }
    \label{fig:ablation}
\end{figure}

To evaluate the robustness of our proposed approach concerning both the graph size and the temporal span used in its construction, a temporal ablation study was conducted.
In addition to the graph used for experimentation, illustrated in Figure~\ref{fig:graph}, we generated four different graphs, each corresponding to one of the four CC-News snapshots (refer to Table~\ref{tab:graphs} in Section~\ref{subsec:graph} for detailed information on each graph).
Subsequently, employing each of these four graphs, we replicated the evaluation procedure described in Section~\ref{subsec:classification}.
These evaluations allowed us to measure how the performance of the proposed strategies changed, when changing the graph, compared to the reported values in Table~\ref{tab:results}.

Figure~\ref{fig:ablation} shows, without loss of generality, the results obtained with the two best-performing strategies reported in Table~\ref{tab:results}, \textit{P-Reliability} and \textit{I-Reliability}, on the largest experimental set \textit{ExpsetB}.
We observed that, independently of the strategy and the dataset, the best results were always obtained with the largest (in size and time) graph joining all the snapshots.
We also observed that not all strategies exhibit equal robustness to changes in the graph.
For instance, we can see in Figure~\ref{fig:ablation} how \textit{P-Reliability} is more sensitive than \textit{I-Reliability} to the choice of snapshot for graph construction.
Despite this variation, both strategies demonstrate improvement and achieve their best results with reduced uncertainty when considering all snapshots.

\section{Graph Construction Steps}
\label{app:graph-pipeline}

The \textit{Common Crawl News Dataset} (\textit{CC-News})\footnote{\url{https://commoncrawl.org/blog/news-dataset-available}} is published as \textit{WARC} files\footnote{A file format that resembles the raw HTTP request and response of each crawled web page.} grouped by year and month, called snapshots.\footnote{\url{https://data.commoncrawl.org/crawl-data/CC-NEWS/index.html}}
To construct the news media graph, $G=\langle S, E, w\rangle$, from CC-News snapshots, we follow to the subsequent steps:
\begin{enumerate}
    \item Download each WARC file and parse each news article in it to extract its URL and all the hyperlinks in its body. At the end of this step, we have a set of news article URLs, $U$, and a set of hyperlinks $L_u$ for each article $u\in U$.
    \item Generate the graph nodes $S$ from $U$ simply as $S = \{domain(u):u \in U\}$ which contains the URL domain names (\textit{e.g.} ``nytimes.com'', ``cnn.com'', etc.) of all processed news articles.
    \item For each domain $s\in S$ create the list of all its hyperlinks, $L_s$, by aggregating the hyperlinks of all its articles, \textit{i.e.} $L_s = \bigcup_{u \in U : domain(u) = s} L_u$.
    \item Finally, generate the graph edges $(s, s')\in E$ for each $s\in S$ by creating an edge to each unique domain $s'$ in its hyperlinks $L_s$ weighted by the proportion of links whose domain is $s'$, \textit{i.e.} $w(s, s') = |\{l \in L_s : domain(l) = s'\}| / |L_s|$.
\end{enumerate}

\section{NewsGuard Score Details}
\label{app:newsguard}

In Section~\ref{subsec:correlation}, we used the scores from the \textit{NewsGuard} dataset introduced in Section~\ref{subsec:datasets} to measure the correlation with estimated $\rho(s)$ values.
NewsGuard employs a team of journalists and experienced editors to produce these reliability scores for news and information websites.
The score ranges from 0 to 100 and NewsGuard is transparent about the methodology used to compute it.
Namely, they compute this reliability score based on the following nine apolitical criteria, each is worth the indicated number of points, based on importance:
\begin{enumerate}
    \item \textit{\textbf{Does not repeatedly publish false content?} \textit{(22 points)}}
    \item \textit{\textbf{Gathers and presents information responsibly?} \textit{(18 points)}}
    \item \textit{\textbf{Regularly corrects or clarifies errors?} \textit{(12.5 points)}}
    \item \textit{\textbf{Handles the difference between news and opinion responsibly?} \textit{(12.5 points)}}
    \item \textit{\textbf{Avoids deceptive headlines?} \textit{(10 points)}}
    \item \textit{\textbf{Website discloses ownership and financing?} \textit{(7.5 points)}}
    \item \textit{\textbf{Clearly labels advertising?} \textit{(7.5 points)}}
    \item \textit{\textbf{Reveals who’s in charge, including any possible conflicts of interest?} \textit{(5 points)}}
    \item \textit{\textbf{Provides information about content creators?} \textit{(5 points)}}
\end{enumerate}
More details can be found in the ``Rating Process and Criteria'' section of their website.\footnote{\url{https://www.newsguardtech.com/ratings/rating-process-criteria/}}



\section{Detailed Results}
\label{app:results}

\begin{figure}[t]
    \centering
    \includegraphics[width=.9\linewidth]{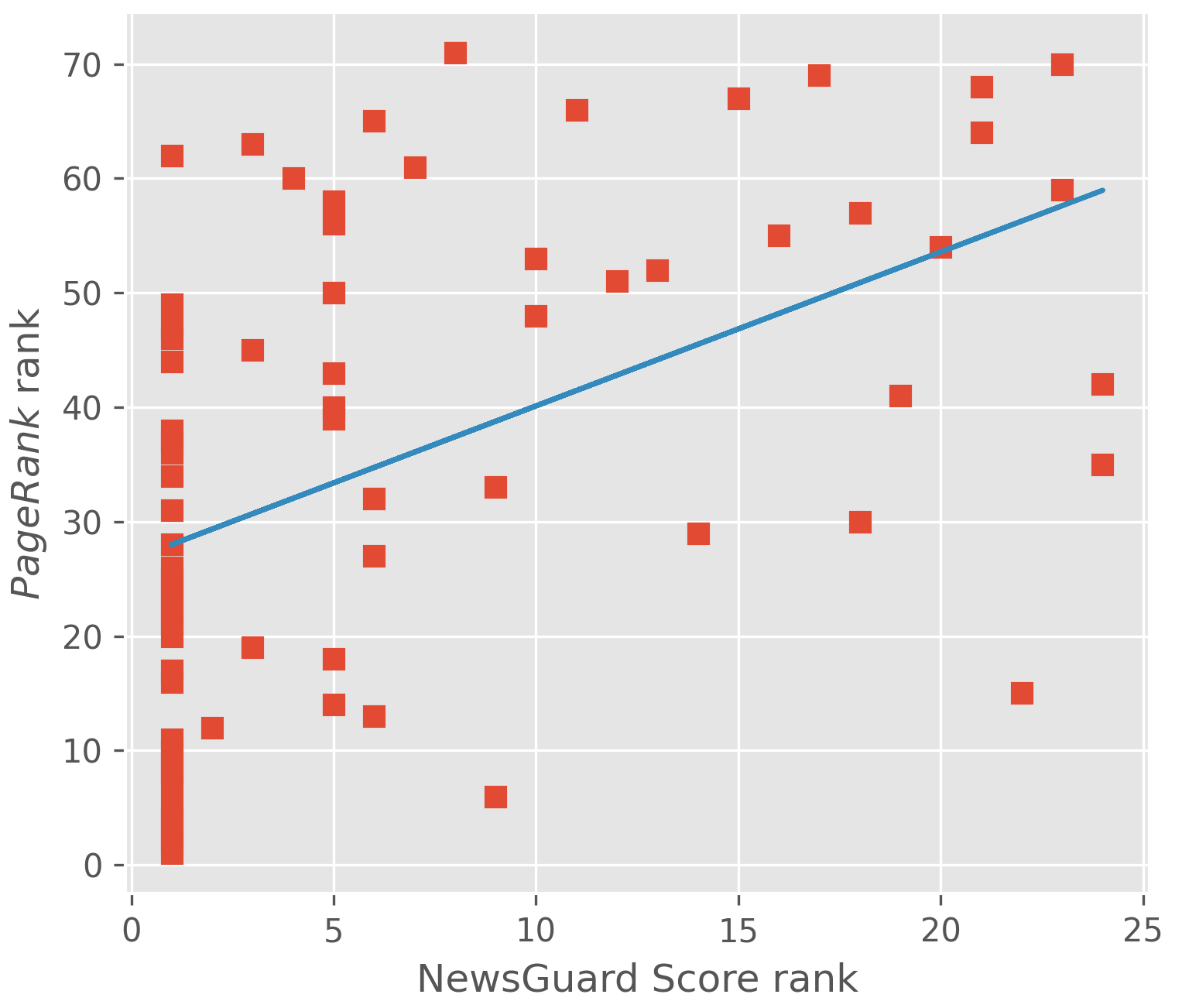}
    \caption{Scatter plot showing the correlation between the rankings obtained by PageRank values (y-axis) and News Guard scores (x-axis).}
    \label{fig:corr-pagerank}
\end{figure}

Table~\ref{tab:results-full} shows the detailed evaluation results obtained on the three experimentation sets along with two naive baselines for reference, random and majority class classifiers.
Furthermore, for \textit{ExpsetA}, we also report the results obtained using the classification models introduced in previous works~\citep{baly-etal-2018-predicting, baly-etal-2020-written} along with the ensemble results.
We can see that on \textit{ExpsetA}, unlike on the other experimental sets, our strategies have 100\% precision for the \textit{unreliable} sources but its recall on the same sources is quite low (from 22.20\% to 38.72\%) showing the models are detecting only a small portion of unreliable sources but with high precision (probably only the easiest cases).
By performing the ensemble with Baly models, the recall for the \textit{unreliable} group increases allowing the graph-based strategies to identify more unreliable sources, in turn improving the macro average F$_1$ scores.

\begin{figure*}[t]
    \centering
    \includegraphics[width=.9\linewidth]{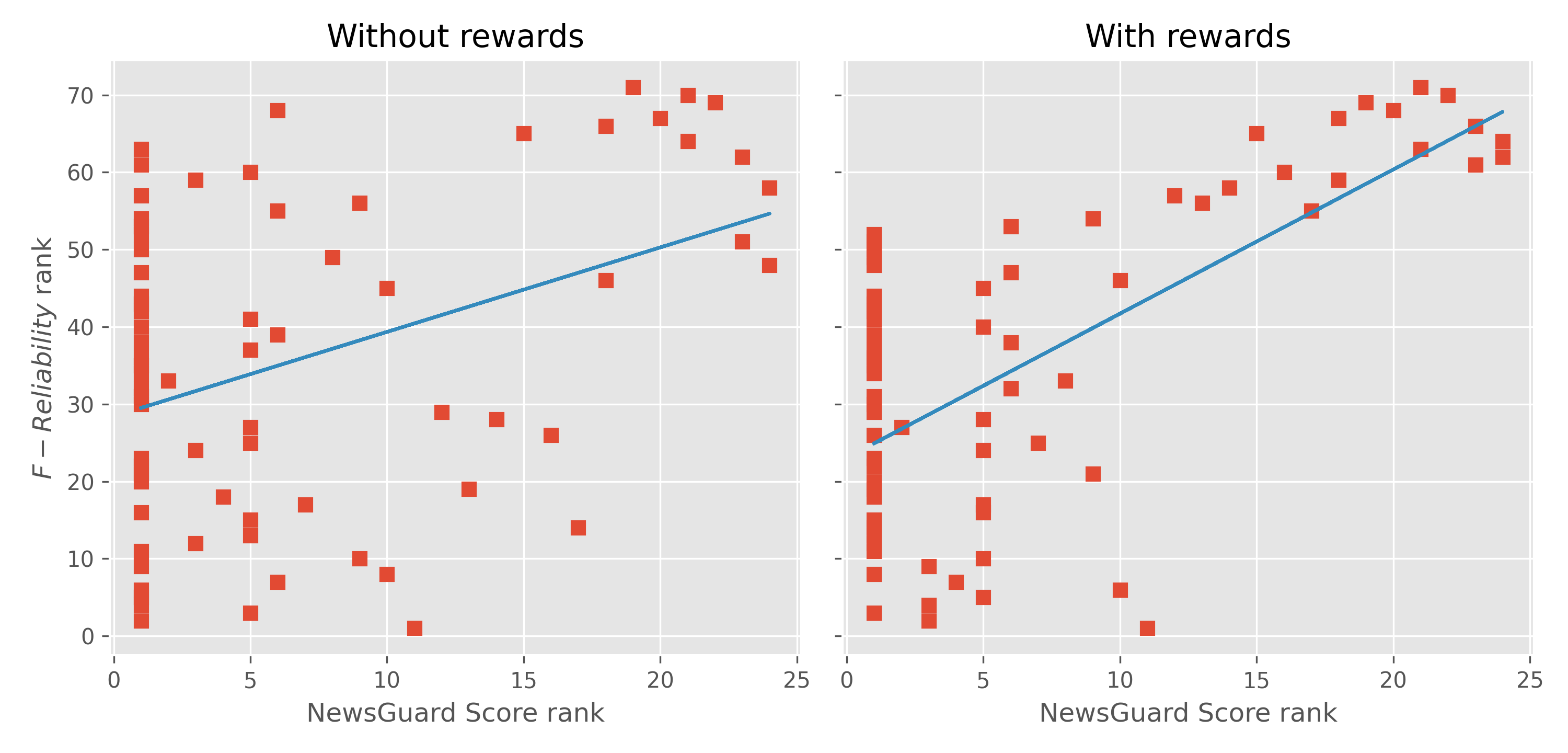}
    \caption{Scatter plot showing the correlation between the rankings obtained by \textit{F-Reliability} values (y-axis) and News Guard scores (x-axis). Left side without rewards and right side with rewards.}
    \label{fig:corr-f}
\end{figure*}

\begin{figure*}[t]
    \centering
    \includegraphics[width=.9\linewidth]{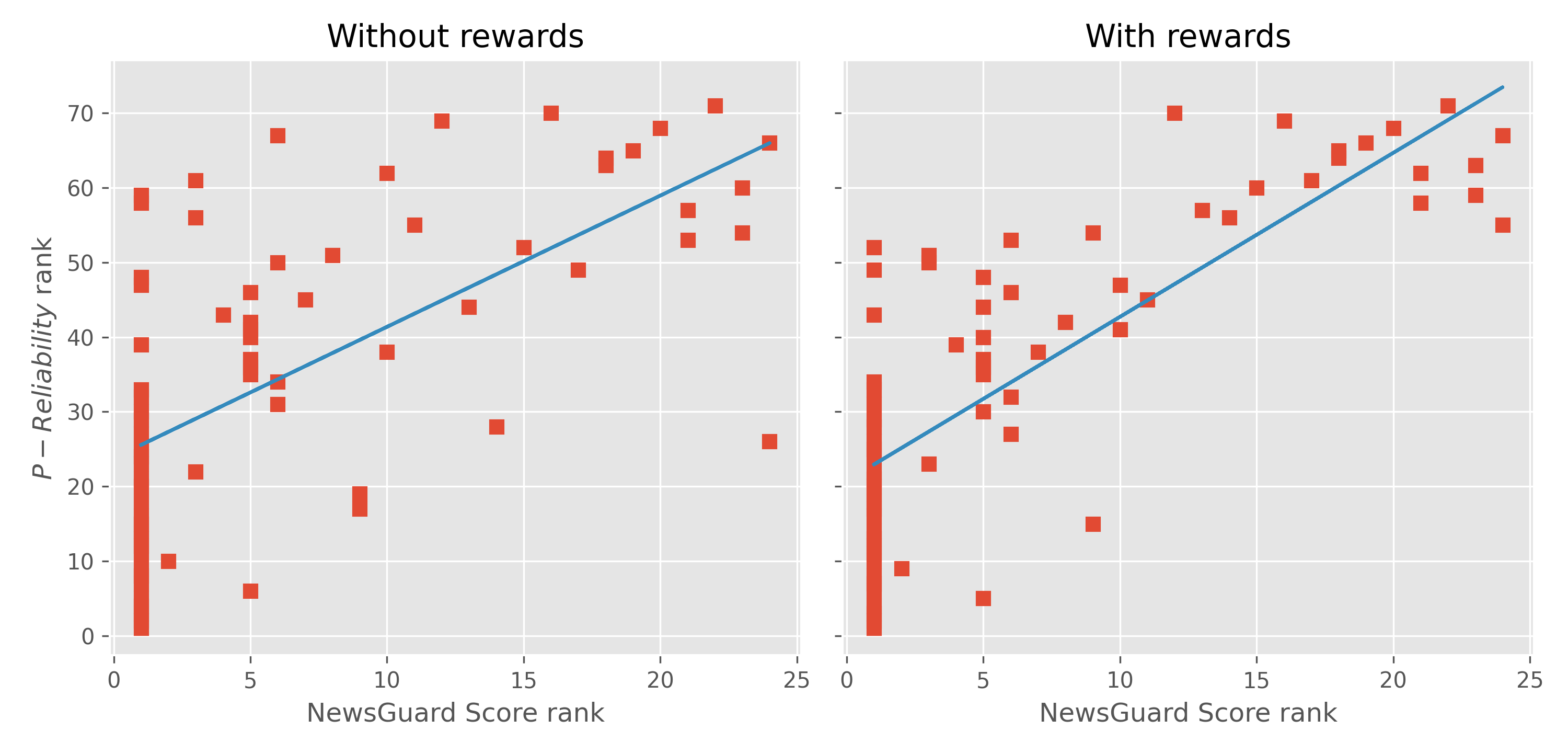}
    \caption{Scatter plot showing the correlation between the rankings obtained by \textit{P-Reliability} values (y-axis) and News Guard scores (x-axis). Left side without rewards and right side with rewards.}
    \label{fig:corr-p}
\end{figure*}

\begin{figure*}[t]
    \centering
    \includegraphics[width=.9\linewidth]{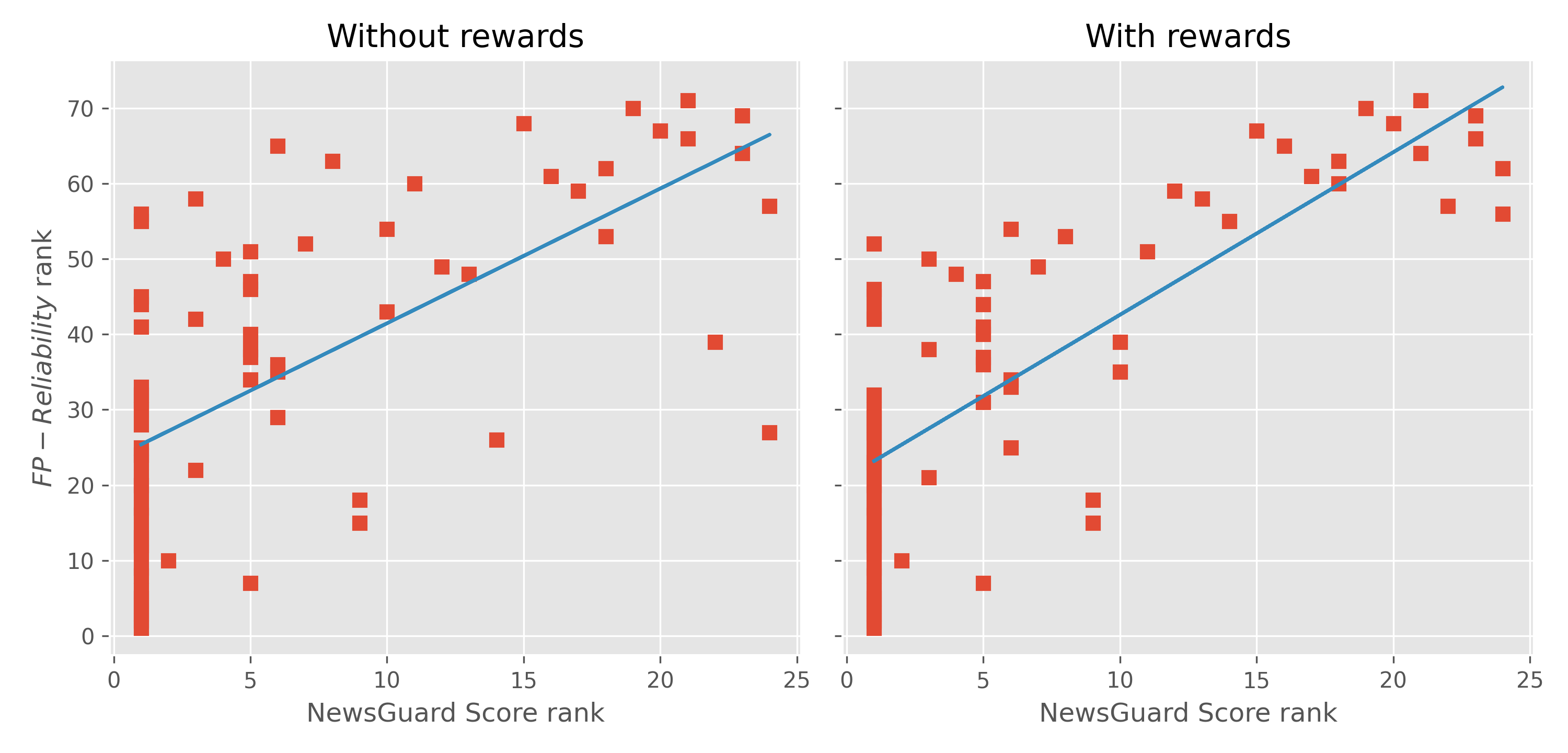}
    \caption{Scatter plot showing the correlation between the rankings obtained by \textit{FP-Reliability} values (y-axis) and News Guard scores (x-axis). Left side without rewards and right side with rewards.}
    \label{fig:corr-fp}
\end{figure*}

\begin{figure*}[t]
    \centering
    \includegraphics[width=.9\linewidth]{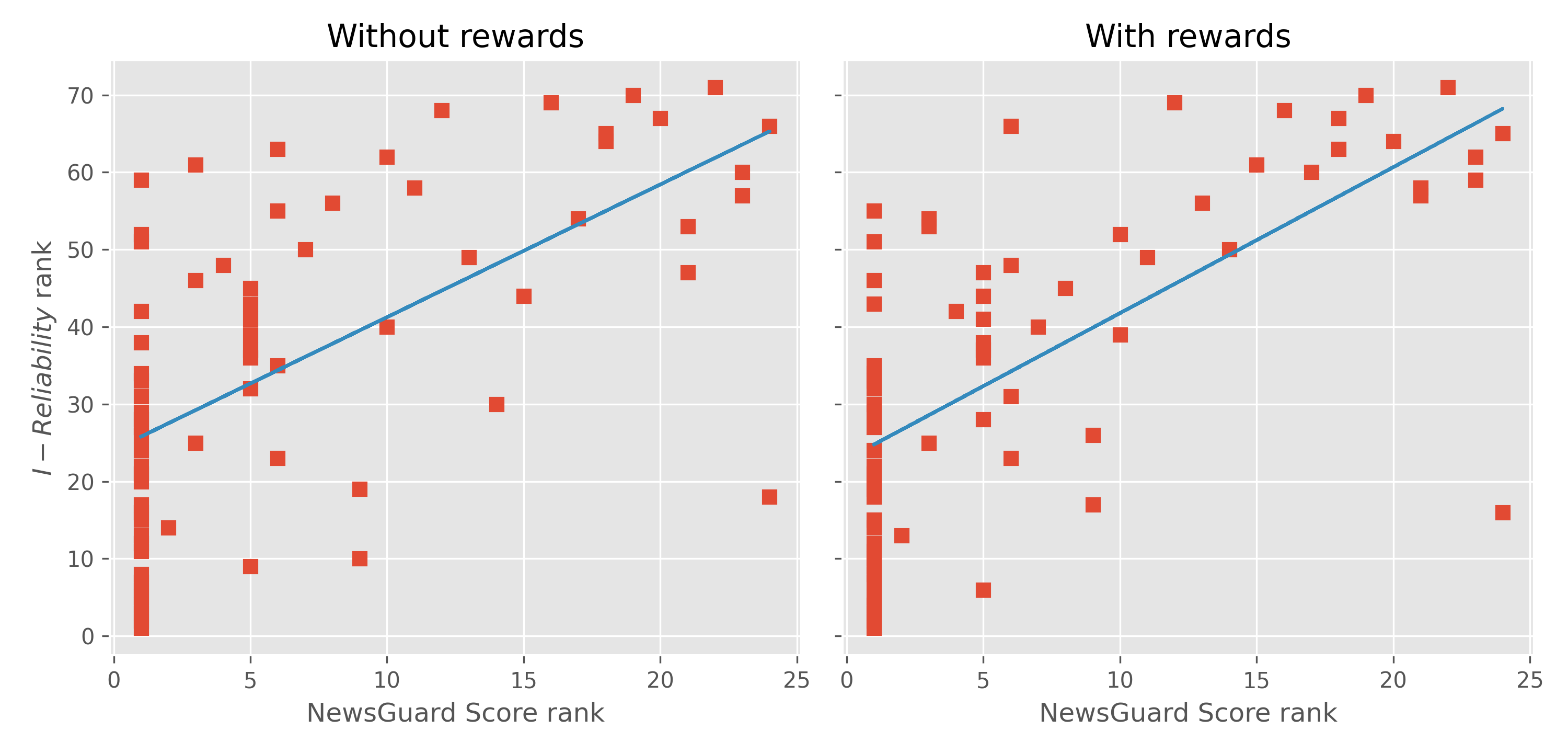}
    \caption{Scatter plot showing the correlation between the rankings obtained by \textit{I-Reliability} values (y-axis) and News Guard scores (x-axis). Left side without rewards and right side with rewards.}
    \label{fig:corr-i}
\end{figure*}

Finally, in Figure~\ref{fig:corr-pagerank} is shown the scatter plot showing the correlation between the ranking obtained by PageRank values and the one obtained by the News Guard scores. Likewise, Figures~\ref{fig:corr-f}, \ref{fig:corr-p}, \ref{fig:corr-fp}, and \ref{fig:corr-i} show the scatter plots for \textit{F-Reliability}, \textit{P-Reliability}, \textit{FP-Reliability}, \textit{I-Reliability}, respectivelly.

\begin{landscape}
\begin{table}
    \centering
    \small
    \begin{tabular}{>{\rowmac}c>{\rowmac}c>{\rowmac}c>{\rowmac}c>{\rowmac}c|>{\rowmac}c>{\rowmac}c>{\rowmac}c|>{\rowmac}c>{\rowmac}c>{\rowmac}c|>{\rowmac}c<{\clearrow}}
        \toprule
        \multirow{2}{*}{\textbf{Data}}&\multirow{2}{*}{\textbf{Strategy}} & \multicolumn{3}{c}{\textbf{Precision}} & \multicolumn{3}{c}{\textbf{Recall}} & \multicolumn{3}{c}{\textbf{F$_1$ score}} &  \\
        \cmidrule(lr){3-11}
        &  & \textit{macro avg.} & \textit{reliable} & \textit{unreliable} & \textit{macro avg.} & \textit{reliable} & \textit{unreliable} & \textit{macro avg.} & \textit{reliable} & \textit{unreliable} & \textbf{Acc.} \\
        \midrule

        \multirow{16}{*}{\rotatebox{90}{\textbf{ExpsetA}}} \setrow{\itshape} & M-BL & 36.72$\pm$1.29 & 73.44$\pm$2.57 & 0.00$\pm$0.00 & 50.00$\pm$0.00 & \underline{\textbf{100.00}}$\pm$0.00 & 0.00$\pm$0.00 & 42.33$\pm$0.85 & 84.66$\pm$1.70 & 0.00$\pm$0.00 & 73.44$\pm$2.57 \\
        \setrow{\itshape} & R-BL & 51.37$\pm$4.21 & 74.63$\pm$3.49 & 28.11$\pm$6.08 & 51.56$\pm$5.30 & 52.87$\pm$6.72 & 50.26$\pm$7.86 & 48.85$\pm$5.29 & 61.76$\pm$5.63 & 35.94$\pm$6.94 & 52.33$\pm$5.52 \\
        \setrow{\itshape} & Baly18 & 70.11$\pm$5.74 & \textbf{82.48}$\pm$5.02 & 57.74$\pm$8.72 & 67.68$\pm$7.52 & 87.65$\pm$4.31 & 47.70$\pm$16.75 & 67.87$\pm$6.89 & 84.81$\pm$2.65 & 50.92$\pm$11.97 & 76.95$\pm$4.11\\
        \setrow{\itshape} & Baly20 & 71.03$\pm$9.32 & 78.42$\pm$3.14 & 63.64$\pm$20.83 & 64.64$\pm$3.62 & 88.74$\pm$7.19 & 40.55$\pm$8.33 & 65.24$\pm$4.03 & 82.99$\pm$2.06 & 47.50$\pm$6.75 & 74.37$\pm$2.84\\
        \cmidrule(lr){3-12}
        & F-R & 89.02$\pm$1.75 & 78.03$\pm$3.51 & \underline{\textbf{100.00}}$\pm$0.00 & 61.10$\pm$4.70 & \underline{\textbf{100.00}}$\pm$0.00 & 22.20$\pm$9.40 & 61.52$\pm$6.83 & 87.62$\pm$2.19 & 35.42$\pm$11.92 & 79.26$\pm$3.59 \\
        & P-R & \underline{\textbf{90.96}}$\pm$0.88 & \textbf{81.91}$\pm$1.75 & \underline{\textbf{100.00}}$\pm$0.00 & \textbf{69.30}$\pm$3.53 & \underline{\textbf{100.00}}$\pm$0.00 & \textbf{38.59}$\pm$7.05 & \textbf{72.67}$\pm$4.10 & \underline{\textbf{90.05}}$\pm$1.05 & \textbf{55.29}$\pm$7.87 & \textbf{83.79}$\pm$1.59 \\
        & FP-R & 90.29$\pm$1.07 & 80.57$\pm$2.14 & \underline{\textbf{100.00}}$\pm$0.00 & 66.53$\pm$3.31 & \underline{\textbf{100.00}}$\pm$0.00 & 33.06$\pm$6.61 & 69.28$\pm$3.91 & 89.23$\pm$1.32 & 49.34$\pm$7.13 & 82.29$\pm$2.07 \\
        & I-R & \textbf{90.95}$\pm$1.02 & \textbf{81.89}$\pm$2.04 & \underline{\textbf{100.00}}$\pm$0.00 & \textbf{69.36}$\pm$2.75 & \underline{\textbf{100.00}}$\pm$0.00 & \textbf{38.72}$\pm$5.51 & \textbf{72.81}$\pm$3.18 & \textbf{90.03}$\pm$1.23 & \textbf{55.60}$\pm$5.86 & \textbf{83.77}$\pm$1.81 \\

        \cmidrule(lr){3-12}

        & \textit{F-R+Baly18} & 76.54$\pm$3.69 & 87.12$\pm$2.64 & 65.95$\pm$7.81 & 75.85$\pm$3.02 & 87.65$\pm$4.31 & 64.06$\pm$7.47 & 75.84$\pm$3.01 & 87.29$\pm$2.02 & 64.39$\pm$4.59 & 81.32$\pm$2.67 \\
        & \textit{F-R+Baly20} & \textbf{77.80}$\pm$7.68 & 84.57$\pm$2.43 & \textbf{71.04}$\pm$17.10 & 74.79$\pm$2.98 & \textbf{88.74}$\pm$7.19 & 60.84$\pm$3.90 & 75.34$\pm$4.18 & 86.38$\pm$2.69 & 64.30$\pm$5.75 & 80.30$\pm$3.67 \\
        & \textit{P-R+Baly18} & 77.47$\pm$3.96 & \textbf{87.99}$\pm$1.21 & 66.95$\pm$8.47 & \textbf{77.23}$\pm$1.76 & 87.65$\pm$4.31 & \textbf{66.82}$\pm$3.31 & \textbf{77.11}$\pm$2.95 & \textbf{87.75}$\pm$2.06 & \textbf{66.47}$\pm$4.27 & \textbf{82.11}$\pm$2.72 \\
        & \textit{P-R+Baly20} & 77.04$\pm$7.79 & 83.89$\pm$2.25 & \textbf{70.19}$\pm$17.23 & 73.70$\pm$2.95 & \textbf{88.74}$\pm$7.19 & 58.66$\pm$4.34 & 74.36$\pm$4.10 & 86.02$\pm$2.57 & 62.70$\pm$5.84 & 79.69$\pm$3.50 \\
        & \textit{FP-R+Baly18} & 74.25$\pm$4.78 & 85.08$\pm$3.73 & 63.42$\pm$8.54 & 72.67$\pm$4.33 & 87.65$\pm$4.31 & 57.68$\pm$8.86 & 73.01$\pm$4.32 & 86.24$\pm$2.66 & 59.79$\pm$6.25 & 79.54$\pm$3.72 \\
        & \textit{FP-R+Baly20} & 76.01$\pm$7.86 & 82.96$\pm$3.47 & 69.06$\pm$17.57 & 72.10$\pm$4.12 & \textbf{88.74}$\pm$7.19 & 55.46$\pm$9.83 & 72.67$\pm$4.28 & 85.46$\pm$2.11 & 59.88$\pm$7.00 & 78.72$\pm$2.99 \\
        & \textit{I-R+Baly18} & \textbf{77.79}$\pm$3.45 & \underline{\textbf{88.29}}$\pm$1.75 & 67.30$\pm$7.86 & \underline{\textbf{77.73}}$\pm$1.47 & 87.65$\pm$4.31 & \underline{\textbf{67.82}}$\pm$4.62 & \underline{\textbf{77.47}}$\pm$2.40 & \textbf{87.89}$\pm$1.89 & \underline{\textbf{67.06}}$\pm$3.40 & \textbf{82.34}$\pm$2.40 \\
        & \textit{I-R+Baly20} & 76.12$\pm$7.77 & 82.85$\pm$2.25 & 69.38$\pm$17.58 & 72.10$\pm$2.28 & \textbf{88.74}$\pm$7.19 & 55.46$\pm$3.69 & 72.88$\pm$3.46 & 85.46$\pm$2.36 & 60.30$\pm$4.77 & 78.74$\pm$3.13 \\

        \midrule
        \midrule

        \multirow{6}{*}{\rotatebox{90}{\textbf{ExpsetB}}} \setrow{\itshape} & M-BL & 38.50$\pm$0.04 & 77.00$\pm$0.09 & 0.00$\pm$0.00 & 50.00$\pm$0.00 & \underline{\textbf{100.00}}$\pm$0.00 & 0.00$\pm$0.00 & 43.50$\pm$0.03 & 87.00$\pm$0.06 & 0.00$\pm$0.00 & 77.00$\pm$0.09 \\
        \setrow{\itshape} & R-BL & 51.09$\pm$1.39 & 78.05$\pm$1.30 & 24.13$\pm$1.49 & 51.53$\pm$1.96 & 51.72$\pm$3.06 & 51.34$\pm$3.70 & 47.48$\pm$1.83 & 62.17$\pm$2.45 & 32.80$\pm$1.93 & 51.63$\pm$2.21 \\

        \cmidrule(lr){3-12}

        & \textit{F-R} & 61.74$\pm$3.19 & 83.63$\pm$0.71 & 39.85$\pm$5.85 & 63.24$\pm$2.10 & 76.38$\pm$6.35 & 50.11$\pm$2.82 & 61.85$\pm$3.12 & 79.72$\pm$3.69 & 43.98$\pm$2.61 & 70.34$\pm$4.36 \\
        & \textit{P-R} & \textbf{74.66}$\pm$2.51 & \underline{\textbf{88.44}}$\pm$1.09 & \textbf{60.89}$\pm$4.20 & \underline{\textbf{74.78}}$\pm$2.26 & \textbf{88.16}$\pm$1.62 & \textbf{61.41}$\pm$3.72 & \textbf{74.69}$\pm$2.31 & \textbf{88.29}$\pm$1.15 & \textbf{61.10}$\pm$3.52 & \textbf{82.00}$\pm$1.73 \\
        & \textit{FP-R} & 59.01$\pm$1.95 & 85.73$\pm$1.42 & 32.29$\pm$2.67 & 62.60$\pm$2.66 & 56.63$\pm$5.39 & \underline{\textbf{68.57}}$\pm$3.22 & 55.95$\pm$3.21 & 68.08$\pm$4.09 & 43.82$\pm$2.55 & 59.38$\pm$3.98 \\
        & \textit{I-R} & \underline{\textbf{76.68}}$\pm$3.08 & \textbf{87.98}$\pm$1.10 & \underline{\textbf{65.39}}$\pm$5.31 & \textbf{74.60}$\pm$2.40 & \underline{\textbf{90.67}}$\pm$1.68 & 58.53$\pm$3.78 & \underline{\textbf{75.51}}$\pm$2.61 & \underline{\textbf{89.30}}$\pm$1.22 & \underline{\textbf{61.72}}$\pm$4.02 & \underline{\textbf{83.28}}$\pm$1.87 \\

        \midrule
        \midrule

        \multirow{6}{*}{\rotatebox{90}{\textbf{ExpsetB$^-$}}} \setrow{\itshape} & M-BL & 45.38$\pm$0.05 & 90.76$\pm$0.10 & 0.00$\pm$0.00 & 50.00$\pm$0.00 & \underline{\textbf{100.00}}$\pm$0.00 & 0.00$\pm$0.00 & 47.58$\pm$0.03 & 95.15$\pm$0.06 & 0.00$\pm$0.00 & 90.76$\pm$0.10 \\
        \setrow{\itshape} & R-BL & 49.82$\pm$2.41 & 90.62$\pm$2.36 & 9.02$\pm$2.47 & 49.43$\pm$7.19 & 48.34$\pm$0.88 & 50.52$\pm$14.11 & 39.17$\pm$2.54 & 63.04$\pm$1.09 & 15.31$\pm$4.20 & 48.55$\pm$1.71 \\

        \cmidrule(lr){3-12}

        & \textit{F-R} & 60.60$\pm$2.73 & 94.14$\pm$0.58 & 27.07$\pm$5.03 & 66.84$\pm$2.97 & 86.69$\pm$3.07 & 46.99$\pm$5.59 & 62.18$\pm$3.15 & 90.23$\pm$1.75 & 34.12$\pm$4.90 & 83.02$\pm$2.77 \\
        & \textit{P-R} & \textbf{77.75}$\pm$5.20 & \underline{\textbf{96.46}}$\pm$0.83 & \textbf{59.04}$\pm$9.83 & \underline{\textbf{80.48}}$\pm$4.25 & \textbf{95.21}$\pm$1.42 & \textbf{65.74}$\pm$7.79 & \textbf{78.90}$\pm$4.50 & \textbf{95.83}$\pm$0.98 & \textbf{61.97}$\pm$8.04 & \textbf{92.48}$\pm$1.74 \\
        & \textit{FP-R} & 59.20$\pm$0.39 & 95.98$\pm$0.63 & 22.41$\pm$0.49 & 72.21$\pm$1.94 & 75.77$\pm$2.32 & 68.66$\pm$5.99 & 59.20$\pm$0.54 & 84.66$\pm$1.23 & 33.74$\pm$0.83 & 75.11$\pm$1.60 \\
        & \textit{I-R} & \underline{\textbf{83.22}}$\pm$4.09 & \textbf{96.13}$\pm$0.72 & \underline{\textbf{70.32}}$\pm$7.83 & \textbf{79.41}$\pm$3.58 & \underline{\textbf{97.30}}$\pm$0.85 & 61.52$\pm$7.02 & \underline{\textbf{81.05}}$\pm$3.40 & \underline{\textbf{96.71}}$\pm$0.60 & \underline{\textbf{65.39}}$\pm$6.22 & \underline{\textbf{93.99}}$\pm$1.09 \\

        \bottomrule
    \end{tabular}
    \caption{5-fold cross-validation detailed results for reliability classification. Mean and standard deviation is shown for each metric.
    \textbf{Bold} indicates 2nd-best-performing values, while \underline{\textbf{underlined}} the best-performing values in each dataset.
    R-BL and M-BL refer to random and majority class baselines; Baly18 and Baly20 refer to \citet{baly-etal-2018-predicting} and \citet{baly-etal-2020-written}; and \textit{*}-R stands for \textit{*}-Reliability.
    }
    \label{tab:results-full}
\end{table}
\end{landscape}

\end{document}